\documentclass{article} 
\usepackage{colm}
\usepackage[colorlinks = true,
            linkcolor = blue,
            urlcolor  = blue,
            citecolor = blue,
            anchorcolor = blue]{hyperref}   
\usepackage[utf8]{inputenc} 
\usepackage[T1]{fontenc}    
\usepackage{url}            
\usepackage{booktabs}       
\usepackage{amsfonts}       
\usepackage{nicefrac}       
\usepackage{microtype}      
\usepackage{xcolor}         

\usepackage{booktabs}
\usepackage{multirow}
\usepackage{tabularx}
\usepackage{makecell}
\usepackage{amssymb}
\usepackage{pifont}
\usepackage{amsmath}
\usepackage{tcolorbox}
\usepackage{listings}
\usepackage{xcolor}
\tcbuselibrary{skins}
\usepackage{hyperref}
\usepackage{float}
\usepackage{tcolorbox}
\tcbuselibrary{breakable}

\usepackage{fontawesome}

\usepackage{bbding}
\makeatletter
  \newcommand\figcaption{\def\@captype{figure}\caption}
  \newcommand\tabcaption{\def\@captype{table}\caption}
\makeatother

\usepackage[utf8]{inputenc} 
\usepackage[T1]{fontenc}    
\usepackage{url}            
\usepackage{booktabs}       
\usepackage{amsfonts}       
\usepackage{nicefrac}       
\usepackage{microtype}      
\usepackage{array}
\newcolumntype{C}[1]{>{\centering\arraybackslash}m{#1}}
\usepackage{xcolor}         
\usepackage{color, colortbl}
\definecolor{citecolor}{HTML}{2980b9}
\definecolor{linkcolor}{HTML}{c0392b}
\definecolor{darkorange}{HTML}{FF8C00}
\definecolor{chocolate}{HTML}{D2691E}
\definecolor{darkgreen}{HTML}{006400}
\definecolor{darkblue}{HTML}{00008B}
\definecolor{mediumblue}{HTML}{0000CD}
\definecolor{dodgerblue}{HTML}{1E90FF}
\definecolor{royalblue}{HTML}{4169E1}
\definecolor{shadecolor}{RGB}{237,237,237}
\definecolor{backred}{RGB}{255, 190, 190}
\definecolor{backblue}{RGB}{210, 230, 250}
\usepackage{colortbl}
\definecolor{graybg}{gray}{0.9}
\definecolor{zrrgreen}{HTML}{008000}
\definecolor{zrrblue}{HTML}{4682B4}
\definecolor{zrrred}{HTML}{B22222}
\definecolor{purple1}{RGB}{126, 107, 196}
\definecolor{purple2}{RGB}{199, 158, 207}
\definecolor{purple3}{RGB}{214, 200, 255}
\definecolor{purple4}{RGB}{254, 240, 255}
\usepackage{amsmath}
\usepackage{amssymb}
\usepackage{graphicx}

\usepackage{multirow}
\usepackage{makecell}
\usepackage{caption}

\usepackage{adjustbox}
\usepackage{wrapfig}

\usepackage{float}
\usepackage{subfig}
\usepackage[linesnumbered,ruled,vlined]{algorithm2e}

\usepackage{fancyvrb}
\usepackage{fvextra}
\usepackage{xcolor}
\usepackage{caption}

\tcbset{
  promptbox/.style={
    enhanced,
    breakable,
    colback=gray!3,
    colframe=gray!60,
    boxrule=0.5pt,
    arc=2pt,
    left=4pt,
    right=4pt,
    top=4pt,
    bottom=4pt,
    fonttitle=\bfseries,
    coltitle=black,
    colbacktitle=gray!50,
    extras middle and last={
      borderline north={0pt}{0pt}{white}
    },
    extras first and middle={
      borderline south={0pt}{0pt}{white}
    }
  }
}

\DefineVerbatimEnvironment{PromptVerb}{Verbatim}{
  fontsize=\scriptsize,
  breaklines=true,
  breakanywhere=true,
  breaksymbolleft={},
  breaksymbolright={}
}


\usepackage{framed}
\usepackage{makecell}
\usepackage{listings}
\definecolor{lightgray}{rgb}{.9,.9,.9}
\definecolor{darkgray}{rgb}{.4,.4,.4}
\definecolor{purple}{rgb}{0.65, 0.12, 0.82}
\lstdefinelanguage{JavaScript}{
  keywords={break, case, catch, continue, debugger, default, delete, do, else, false, finally, for, function, if, in, instanceof, new, null, return, switch, this, throw, true, try, typeof, var, void, while, with},
  morecomment=[l]{//},
  morecomment=[s]{/*}{*/},
  morestring=[b]',
  morestring=[b]",
  ndkeywords={class, export, boolean, throw, implements, import, this},
  keywordstyle=\color{blue}\bfseries,
  ndkeywordstyle=\color{darkgray}\bfseries,
  identifierstyle=\color{black},
  commentstyle=\color{purple}\ttfamily,
  stringstyle=\color{red}\ttfamily,
  sensitive=true
}
\usepackage{xspace}
\usepackage[shortlabels]{enumitem}

\newcommand\blfootnote[1]{%
  \begingroup
  \renewcommand\thefootnote{}\footnote{#1}%
  \addtocounter{footnote}{-1}%
  \endgroup
}

\title{ArenaFlow: From Trajectory Ranking to Hierarchical Credit Propagation for Open-Ended Agent RL}

\author{Qiang Zhang$^{1}$, Ruixue Ding$^{1}$$^{\ast}$, Fanrui Zhang$^{1}$, Xi Chen$^{1}$, Boli Chen$^{1}$, Shihang Wang$^{1}$\\ \textbf{Yinfeng Huang$^{2}$, Yi Zheng$^{2}$, Pengjun Xie$^{1}$, Kaipeng Zhang$^{1}$, Jiawei Liu$^{\ddagger}$, Zheng-Jun Zha}\\
\\
$^{1}$Alibaba Token Hub, Alibaba Group$^{2}$ Amap, Alibaba Group
}

\colmfinalcopy 

\begin{document}

\maketitle

\blfootnote{$^{\ast}$ Project leader, $^{\ddagger}$ Corresponding author}

\begin{abstract}
Reinforcement learning has substantially improved large language model (LLM) agents in verifiable domains, but remains difficult to apply to open-ended agent tasks, where solutions are diverse and reliable scalar rewards are hard to obtain. Recent pairwise evaluation methods alleviate reward discrimination collapse by replacing pointwise scoring with relative preferences. However, they still compress rich comparative feedback into a single trajectory-level reward, obscuring decisive intermediate steps and preventing successful behaviors from being consolidated into reusable skills. We propose \textbf{ArenaFlow}, a hierarchical credit propagation framework for open-ended agent reinforcement learning. ArenaFlow leverages tournament-based relative ranking to derive trajectory-level reward signals. Each comparison is further equipped with structured reflective evaluation, which reveals three types of supervision: pivotal success steps, reusable strategy skills, and usage attribution of retrieved skills. At the step level, ArenaFlow propagates trajectory-level advantages to high-confidence pivotal steps according to tournament survival depth, enabling more targeted optimization of local reasoning behaviors. At the skill level, ArenaFlow estimates skill utility from group-level usage attribution and maintains a global skill memory through utility-aware updating, pruning, and retrieval. The resulting high-utility skills further serve as policy priors for future exploration. Extensive experiments validate ArenaFlow’s effectiveness on open-ended agent tasks.
The code is available at \href{https://github.com/Alibaba-NLP/qqr}{https://github.com/Alibaba-NLP/qqr}.

\end{abstract}

\section{Introduction}

Large language model (LLM) agents have shown strong potential in complex real-world tasks that require long-horizon reasoning and interaction with external environments \cite{team2025tongyi, liu2025percache}. Reinforcement learning (RL) has become a central paradigm for improving such agents, especially in verifiable domains such as mathematical reasoning and code generation, where rule-based rewards can be reliably obtained \cite{dong2025agentic}. However, extending RL to open-ended agent tasks remains challenging. Tasks such as deep research report generation \cite{li2025webweaver} and personalized travel planning \cite{ning2025deeptravel} typically feature vast solution spaces and involve multiple subjective evaluation rubrics, making it difficult to design reliable reward signals.

\begin{figure}[t]
    \centering
    \includegraphics[width=\linewidth]{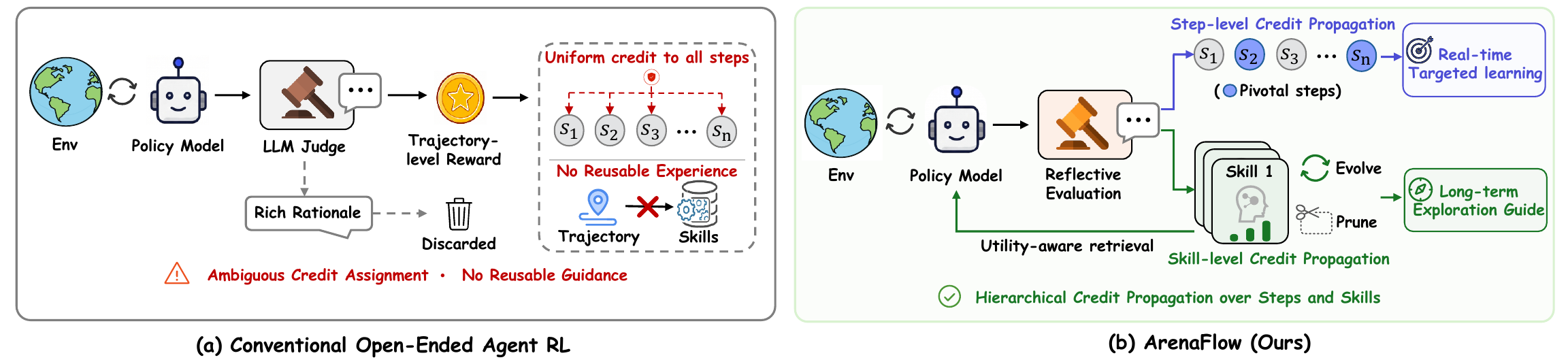}
    \caption{
    \textbf{Comparison between conventional open-ended agent RL and ArenaFlow.} Instead of collapsing judge feedback into a single trajectory-level reward, ArenaFlow propagates comparative feedback to pivotal reasoning steps for targeted policy optimization and to reusable skills for long-term exploration guidance.}  
    \label{fig:first}
    \vspace{-4pt}
\end{figure}

A common solution is to employ LLM-as-a-Judge to evaluate generated trajectories. Early methods often rely on pointwise reward modeling, where each trajectory is assigned an absolute scalar score \cite{viswanathan2025checklists, huang2025reinforcement}. However, such scoring is prone to reward discrimination collapse: when candidate trajectories are generally of high quality, their scores tend to concentrate in a narrow range, causing the optimization signal to be dominated by the intrinsic noise of the judge model \cite{arenarl}. Recent pairwise evaluation methods mitigate this issue by replacing absolute scores with relative preferences, such as reference-based comparison \cite{jia2025writing} or tournament-based ranking \cite{arenarl}. These methods provide more stable supervision by evaluating trajectories in relation to other candidates rather than in isolation. Yet, the resulting preference signal is still exploited primarily at the trajectory level, leaving finer-grained learning signals within and across trajectories largely untapped.

First, a single trajectory-level advantage is typically applied uniformly across a long reasoning trajectory, making it difficult to identify which intermediate reasoning actions truly contribute to success. As a result, decisive local behaviors are diluted by trivial steps. Second, successful reasoning patterns revealed during comparative evaluation are usually discarded after producing the preference decision, rather than being consolidated into reusable experience that can guide future exploration. As illustrated in Figure~\ref{fig:first}, these limitations motivate propagating comparative feedback beyond trajectory-level preferences: locally to pivotal reasoning steps for targeted policy optimization, and across trajectories to consolidate successful behaviors into reusable strategies.

Motivated by this perspective, we propose \textbf{ArenaFlow}, a unified framework for hierarchical credit propagation and skill evolution in open-ended agent RL. ArenaFlow follows the tournament-based relative ranking paradigm, but extends each pairwise comparison with structured reflective evaluation. Specifically, the judge model not only predicts the winning trajectory, but also identifies pivotal success steps, extracts reusable strategy skills, and attributes whether retrieved historical skills are substantively used in decisive reasoning processes. In this way, comparative evaluation yields structured supervision that supports both targeted policy optimization and the accumulation of reusable experience.

ArenaFlow exploits these signals at two complementary levels. First, for step-level credit propagation, ArenaFlow uses tournament survival depth as a confidence indicator: pivotal steps identified in deeper tournament rounds are treated as more reliable contributors to success, since they support victories against stronger candidates. ArenaFlow therefore propagates trajectory-level advantages to these high-confidence pivotal steps through depth-aware weighting and intra-trajectory normalization. Second, for skill-level credit propagation, ArenaFlow estimates the empirical utility of retrieved skills from group-level usage attribution. Skills that are repeatedly used in pivotal success steps receive higher utility, while ineffective or stale skills are pruned from the global memory. During subsequent rollouts, high-utility skills are retrieved as policy priors to guide exploration toward historically successful strategies.

Our contributions are summarized as follows:
\begin{itemize}
    \item We formulate open-ended agent optimization as a hierarchical credit propagation problem, extending trajectory-level preference supervision to pivotal reasoning steps and reusable strategy skills.
    \item We propose tournament-depth step credit propagation, which converts reflective comparison feedback into fine-grained advantages for pivotal reasoning steps while preserving the stability of trajectory-level preference optimization.
    \item We introduce usage-attributed skill evolution, which estimates skill utility from group-level attribution evidence and maintains a global skill memory through utility-aware updating, pruning, and retrieval.
\end{itemize}

\section{Related Work}

\paragraph{Open-Ended Agent RL.}
Reinforcement learning has become a key paradigm for improving the reasoning and tool-use capabilities of LLM agents. In verifiable domains such as mathematical reasoning, methods like PPO \cite{ppo} and GRPO \cite{grpo} can optimize policies effectively with rule-based rewards. However, open-ended agent tasks rarely have a unique correct answer. For example, deep research report generation and personalized travel planning involve subjective, multi-dimensional rubrics, making reliable reward modeling difficult. To address this, recent work increasingly adopts LLM-as-a-Judge for preference-based supervision. Writing-Zero \cite{jia2025writing} assigns binary relative advantages through reference-based comparison. Pref-GRPO \cite{wang2025pref} constructs rewards from cyclic preference evaluation. And ArenaRL \cite{arenarl} formulates optimization as intra-group tournament ranking to improve efficiency and robustness. Unlike these methods, ArenaFlow goes beyond trajectory-level preference optimization by further decomposing comparative feedback into step-level and skill-level credit signals.

\paragraph{Credit Assignment.}
Credit assignment is a core challenge in long-horizon agent RL. Existing methods generally follow two directions \cite{cheng2026stop}. The first uses process reward models to provide denser supervision for intermediate reasoning steps. For example, AgentPRM \cite{agentprm} derives process-level rewards through temporal difference and dependency estimation, while Turn-PPO \cite{Turn-ppo} reformulates multi-turn interactions as turn-level Markov decision processes for turn-level advantage estimation. The second direction focuses on fine-grained credit propagation in structured rollouts. Tree-GRPO \cite{treegrpo} builds tree-structured reasoning expansions to estimate hierarchical relative advantages across branches. GiGPO \cite{gigpo} groups actions under shared anchored states and estimates state-conditioned relative advantages across trajectories. However, in open-ended agent tasks, both intermediate steps and final outputs are often hard to verify objectively, limiting the applicability of existing methods.

\paragraph{Skill Evolution for Agents.}
Constructing reusable skills is an important direction for improving agents' long-term exploration. Early memory-based agents stored raw interaction trajectories as experience replay, but such memories are often verbose and difficult to reuse efficiently \cite{wang2025agent}. Recent methods instead distill compact skills or reflections from historical interactions. Voyager \cite{voyager} continuously accumulates executable skills through automatic curriculum learning, while ExpeL \cite{expel} summarizes successful and failed experiences into reusable natural-language reflections. More recent work integrates skill libraries with policy optimization. SkillRL \cite{skillrl} builds a hierarchical skill library through experience distillation and enables recursive co-evolution between skills and policies. GoS \cite{graphskill} models inter-skill dependencies with graph structures to support compositional skill extraction. However, how to identify truly effective skills in open-ended agent tasks and leverage them to guide continual policy evolution remains underexplored.
\section{Methodology: ArenaFlow}
\label{sec:method}

\begin{figure*}[htbp]
    \centering
    \includegraphics[width=1.0\textwidth]{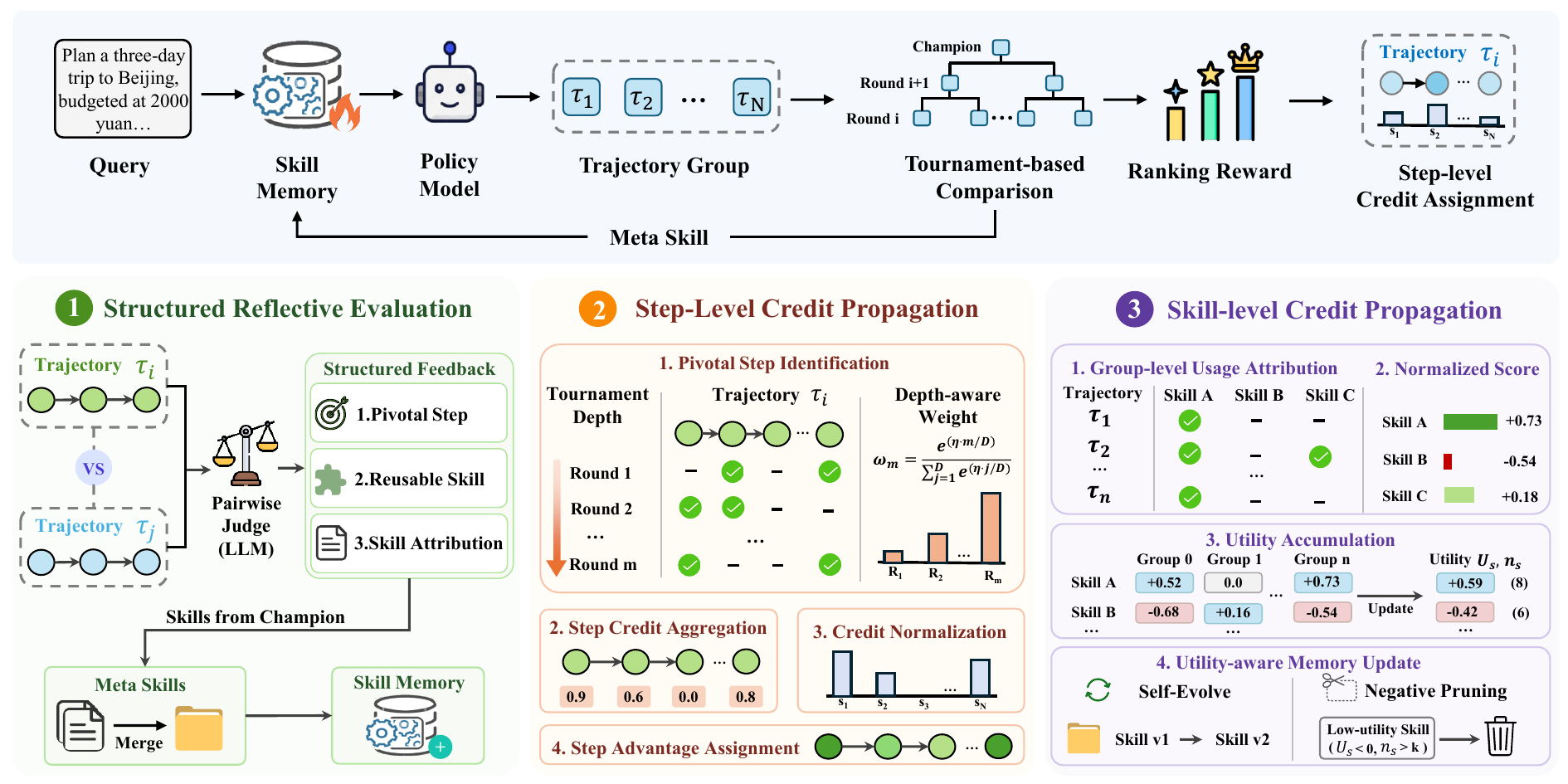}
    \caption{\textbf{Overview of the proposed ArenaFlow.} ArenaFlow first obtains trajectory-level ranking rewards through tournament-based relative comparison, and then propagates the resulting preference signals to two complementary levels.
    At the step level, structured reflective evaluation identifies pivotal reasoning steps and assigns tournament-depth-aware credit for targeted policy optimization.
    At the skill level, usage attribution estimates the empirical utility of retrieved skills, enabling utility-aware memory updating, pruning, and retrieval for future exploration.
}
    \label{fig:method}
\end{figure*}

In this section, we introduce \textbf{ArenaFlow}, a hierarchical credit propagation framework for open-ended agent reinforcement learning.
As illustrated in Figure~\ref{fig:method}, ArenaFlow first obtains robust trajectory-level preference signals through tournament-based relative ranking within each trajectory group.
Instead of applying the resulting trajectory-level advantages uniformly to all reasoning steps, ArenaFlow further propagates these signals to two complementary levels: pivotal reasoning steps for targeted policy optimization, and reusable skills for long-term exploration guidance.

\subsection{Tournament-based Relative Ranking}

To address the inherent discrimination collapse problem in open-ended agent tasks, we follow prior work \cite{arenarl} and employ tournament-based relative ranking to obtain robust trajectory-level preference signals.

Given a user query $x$, the policy model $\pi_\theta$ generates a trajectory group:
\begin{equation}
    \mathcal{G} = \{\tau_1, \tau_2, \ldots, \tau_N\},
\end{equation}
where each trajectory $\tau_i$ consists of a sequence of reasoning and action steps. To obtain high-quality preference signals, we adopt a seeded single-elimination tournament. A greedy decoding trajectory is first used as an anchor to establish an initial ranking prior, after which candidate trajectories are compared through a binary tournament structure.

After the tournament, each trajectory receives an intra-group rank $\mathrm{Rank}_{\tau_i} \in \{0,\ldots,N-1\}$, where rank $0$ denotes the best trajectory. The rank is converted into a quantile-style reward:
\begin{equation}
    r_i = 1 - \frac{\mathrm{Rank}_{\tau_i}}{N-1}.
\end{equation}

The trajectory-level advantage is then obtained by group normalization:
\begin{equation}
    A_i = \frac{r_i - \mu_r}{\sigma_r + \epsilon},
\end{equation}
where $\mu_r$ and $\sigma_r$ denote the mean and standard deviation of rewards within the trajectory group.
ArenaFlow treats $A_i$ as the initial preference signal for hierarchical credit propagation.

\subsection{Structured Reflective Evaluation}

ArenaFlow extends conventional pairwise preference evaluation into structured reflective evaluation. For each pairwise comparison between trajectories $\tau_i$ and $\tau_j$ at tournament round $m$, the judge model $\mathcal{J}$ produces not only a preference decision, but also structured diagnostic feedback.

Specifically, the judge produces three types of reflective information.

\textbf{Pivotal success steps.}
The judge identifies intermediate reasoning and action steps in the winning trajectory that substantially contribute to the pairwise victory. This provides evidence for fine-grained step-level credit propagation.

\textbf{Reusable skills.}
The judge abstracts compact natural-language skills from the compared trajectories.
Each skill summarizes a reusable strategy pattern, including its name, description, core principle, implementation guideline, and failure-avoidance constraint.

\textbf{Skill attribution.}
If prior skills retrieved were injected during rollout, the judge determines whether any of them were substantively used in the pivotal success steps. This attribution is later used to estimate skill utility.

Through structured reflective evaluation, ArenaFlow converts judge rationales into actionable credit signals for both targeted policy optimization and long-term skill evolution.

\subsection{Step-level Credit Propagation}
\label{sec:step_credit}

In long-horizon open-ended trajectories, only a small subset of steps may directly determine final success.
Uniformly assigning the same trajectory-level advantage to all steps can therefore dilute the learning signal of decisive local behaviors.
ArenaFlow addresses this issue by selectively amplifying pivotal success steps identified through structured reflective evaluation.

We use tournament depth as a confidence indicator for step-level credit.
Pivotal steps identified in later tournament rounds are considered more reliable, since they support victories against stronger trajectories that have survived previous comparisons.
Assume the tournament has total depth $D$.
The confidence weight for round $m$ is defined as:
\begin{equation}
    \omega_m
    =
    \frac{
    \exp(\eta \cdot m / D)
    }{
    \sum_{j=1}^{D}
    \exp(\eta \cdot j / D)
    }
\end{equation}
where $\eta$ controls the sharpness of depth weighting.

Let $\mathcal{R}_i$ denote the set of tournament rounds won by trajectory $\tau_i$, and
$\mathcal{P}_{i}^{m}$ denotes the indices of pivotal success steps identified for $\tau_i$ in round $m$. The raw step credit score for the $k$-th step of $\tau_i$ is:
\begin{equation}
    q_{i}^k
    =
    \sum_{m \in \mathcal{R}_i}
    \omega_m
    \cdot
    \mathbb{I}
    \left(
    k \in \mathcal{P}_{i}^{m}
    \right)
\end{equation}
We then normalize the step credit scores within each trajectory:
\begin{equation}
    g_{i}^k
    =
    \frac{
    q_{i}^k
    }{
    \max_{j \in \tau_i} q_{i}^j + \epsilon
    }
\end{equation}
Here, $g_{i}^k$ represents the normalized pivotal-step credit assigned to $k$-th step of $\tau_i$.

Finally, ArenaFlow combines the trajectory-level advantage $A_i$ with the step-level modulation term $g_{i}^k$:
\begin{equation}
    A_{i}^k
    =
    A_i
    +
    \max(A_i,0)
    \cdot
    g_{i}^k
\end{equation}
This design preserves the global preference direction while selectively amplifying pivotal success steps in positively ranked trajectories. In this way, ArenaFlow reinforces locally effective reasoning behaviors, achieving targeted policy optimization.

\subsection{Skill-level Credit Propagation}
\label{sec:skill_utility}

Step-level credit propagation improves local policy learning, but open-ended agents also require reusable prior skills that can guide exploration across tasks.
Therefore, ArenaFlow maintains a global skill memory $\mathcal{B}$, where each skill $s \in \mathcal{B}$ is associated with an empirical utility score $U_s$ and an evaluation count $n_s$.
The skill memory evolves by propagating attribution evidence from tournament victories to reusable skills, which are then selectively retained and retrieved for future rollouts.

\paragraph{Skill Abstraction and Memory Update.}
After tournament-based evaluation, ArenaFlow collects the skills extracted from pairwise comparisons involving the final champion trajectory.
These sub-skills are summarized and integrated by an LLM into a higher-level meta skill, which is then inserted into the global skill memory $\mathcal{B}$.
If the newly generated meta skill is semantically equivalent to an existing entry but provides a more generalizable or fine-grained strategy, ArenaFlow performs version-aware replacement to update the memory without introducing redundant entries.

\paragraph{Group-level Usage Attribution.}
For each query $x$, ArenaFlow retrieves a set of skills
$\mathcal{S}^{\mathrm{ret}}_x \subseteq \mathcal{B}$
and injects them into the system prompt of policy model.
Since all trajectories in the group $\mathcal{G}$ are conditioned on the same retrieved skills, retrieval alone does not imply usefulness.
ArenaFlow therefore assigns utility to a skill only when the structured judge explicitly attributes it to pivotal success steps in pairwise comparisons.

For a trajectory $\tau_i$, let $\mathcal{R}_i$ denote the set of rounds won by $\tau_i$.
For each winning round $m \in \mathcal{R}_i$, the structured judge returns a set of retrieved skills
$\mathcal{Q}_{i}^{m} \subseteq \mathcal{S}^{\mathrm{ret}}_x$
that are substantively used in the pivotal success steps of $\tau_i$.
The group-level attribution credit for a retrieved skill $s$ is defined as:
\begin{equation}
\label{eq:skill_evidence}
C_s^x
=
\sum_{\tau_i \in \mathcal{G}}
\sum_{m \in \mathcal{R}_i}
\omega_m
\cdot
\mathbb{I}
\left(
s \in \mathcal{Q}_{i}^{m}
\right),
\end{equation}
where $\omega_m$ is the tournament-depth weight defined in Section~\ref{sec:step_credit}.
Thus, a skill receives stronger evidence when it is repeatedly attributed to pivotal success steps, especially in deeper tournament rounds.

To convert this raw evidence into a comparable utility signal, ArenaFlow normalizes $C_s^x$ among all retrieved skills:
\begin{equation}
\label{eq:skill_signal}
z_s^x
=
\frac{
C_s^x-\mu_C^x
}{
\max_{s' \in \mathcal{S}^{\mathrm{ret}}_x}
\left|C_{s'}^x-\mu_C^x\right|
+\epsilon
},
\end{equation}
where $\mu_C^x$ denote the mean of skill credit within the retrieved skill set $\mathcal{S}^{\mathrm{ret}}_x$.


\paragraph{Utility Estimation and Pruning.}
For each retrieved skill $s$, ArenaFlow maintains an online estimate of its long-term utility. 
After obtaining the current attribution-based signal $z_s^x$, the utility estimate is updated as
\begin{equation}
    U_s
    \leftarrow
    \frac{
    n_s U_s + z_s^x
    }{
    n_s + 1
    },
    \qquad
    n_s
    \leftarrow
    n_s + 1 .
\end{equation}
where $n_s$ denotes the number of previous evaluations of skill $s$. 

To prevent stale or misleading skills from affecting future exploration, ArenaFlow maintains an active retrieval pool:
$\mathcal{B}_{\mathrm{active}}$:
\begin{equation}
\label{eq:11}
\mathcal{B}_{\mathrm{active}}
=
\left\{
s \in \mathcal{B}
\;\middle|\;
n_s < K
\;\lor\;
U_s \ge 0
\right\}
\end{equation}
where $K$ is the minimum number of evaluations before pruning.
Equivalently, a skill is removed from the active pool only if it has been evaluated at least $K$ times and still exhibits negative utility.
This warm-up mechanism avoids prematurely discarding newly acquired skills while keeping the active memory compact and reliable.

\subsection{Utility-Aware Skill Retrieval}

ArenaFlow reuses the global skill memory by retrieving skills as strategic priors for future rollouts. 
A desirable retrieval mechanism should select skills that are both semantically relevant to the current query and empirically useful according to historical attribution evidence. 
To this end, ArenaFlow performs retrieval in two stages.

Given a query $x$, ArenaFlow first conducts semantic recall over the active skill memory 
$\mathcal{B}_{\mathrm{active}}$ and obtains the top-$2k$ candidate skills according to cosine similarity $\mathrm{Sim}(x,s)$. 
This stage favors coverage and preserves potentially relevant skills. 
The recalled candidates are then reranked by a utility-aware score:
\begin{equation}
\label{eq:12}
    \phi_{\mathrm{ret}}(x,s)
    =
    \mathrm{Sim}(x,s)
    +
    \lambda \cdot U_s
\end{equation}
where $\lambda$ controls the balance between query relevance and empirical usefulness. 
Finally, the top-$k$ skills are inserted into the system prompt to guide trajectory generation.

\subsection{Policy Optimization}
\label{sec:optimization}

ArenaFlow optimizes the policy with the step-level advantages obtained from hierarchical credit propagation. For trajectory $\tau_i=(\tau_{i}^1,\ldots,\tau_{i}^{L_i})$, the final advantage assigned to step $\tau_{i}^k$ is $A_{i}^k$, which combines the trajectory-level preference signal with the depth-weighted pivotal step credit.

The policy is optimized by maximizing the following objective function:

\begin{equation}
\max_{\pi_\theta}
\mathbb{E}_{x,\mathcal{G}}
\left[
\sum_{\tau_i \in \mathcal{G}}
\sum_{k=1}^{L_i}
\min
\left(
\rho_{i}^k A_{i}^k,
\mathrm{clip}(\rho_{i}^k,1-\epsilon,1+\epsilon) A_{i}^k
\right)
\right]
\end{equation}
where \(\rho_i^k\) denotes the probability ratio. This objective preserves the stability of trajectory-level preference optimization while assigning stronger learning signals to pivotal success steps. Meanwhile, the global skill memory evolves from usage-attributed feedback and provides high-utility strategy priors for future exploration. 

\section{Experiments}
\subsection{Experimental Settings}

\paragraph{Benchmarks.}
We evaluate ArenaFlow on three open-ended agent benchmarks. \textbf{Open-Travel}~\cite{arenarl}, which covers personalized itinerary planning across routing, daily planning, transportation comparison, POI search, and multi-day planning; \textbf{Open-DeepResearch}~\cite{arenarl}, which evaluates multi-turn web search and information synthesis, for research reports and concept explanation; and \textbf{DeepResearch Bench}~\cite{deepresearch_bench}, which contains 100 PhD-level research tasks across 22 domains. 
Details are provided in Appendix~\ref{appendix:benchmark}.

\paragraph{Baselines.}
We compare ArenaFlow with three groups of baselines. 
(1) Strong closed-source models, including GPT-5.2~\cite{gpt}, Grok-4~\cite{grok}, Gemini-2.5-Pro~\cite{gemini}, and Claude-4.5-Sonnet~\cite{claude}. 
(2) Representative RL algorithms, including pointwise reward methods such as GRPO~\cite{grpo}, GSPO~\cite{gspo}, and Reinforce++~\cite{reinforce++}, as well as methods based on pairwise preferences including Writing-Zero~\cite{jia2025writing}, Pref-GRPO~\cite{wang2025pref}, and ArenaRL~\cite{arenarl}. 
ArenaRL is trained directly with RL, while the other RL baselines follow a cold-start-then-RL pipeline.
For a fair comparison, all RL baselines use the same backbone model and judge model as ArenaFlow. 
(3) Proprietary deepsearch systems, including Grok Deeper Search~\cite{grok} and Perplexity Deep Research~\cite{perplexity2026sonardeepresearch}.
Details are provided in Appendix~\ref{appendix:baseline}.

\paragraph{Implementation Details.}
We use Qwen3-8B~\cite{yang2025qwen3} as the backbone model for ArenaFlow and all trainable baselines. 
The RL experiments are optimized with Adam using a learning rate of $1 \times 10^{-6}$ for 120 training steps.
ArenaFlow uses Qwen3-Max~\cite{yang2025qwen3} as the judge for structured reflective evaluation. 
The group size is set to $N=16$, and each training step contains $G=8$ groups. 
For skill retrieval, ArenaFlow retrieves the top-$k=3$ skills. 
The minimum evaluation count for skill pruning in Eq.~\ref{eq:11} is set to $5$, and the utility-aware reranking coefficient in Eq.~\ref{eq:12} is set to $0.2$. 
All experiments are conducted on 8 NVIDIA H20 GPUs.

\subsection{Main Results}

\begin{table*}[t]
\centering
\footnotesize
\setlength{\tabcolsep}{3.4pt}
\caption{Performance comparison on Open-Travel and Open-DeepResearch benchmarks.}
\label{tab:main}
\resizebox{1\textwidth}{!}{
\begin{tabular}{l @{\hspace{6pt}} ccccc c @{\hspace{6pt}} ccccccc c @{\hspace{6pt}} c}
\toprule
\multirow{2}{*}{\textbf{Method}} 
& \multicolumn{6}{c}{\textbf{Open-Travel}} 
& \multicolumn{8}{c}{\textbf{Open-DeepResearch}} 
& \multirow{2}{*}{\textbf{Overall}} \\
\cmidrule(r){2-7} \cmidrule(lr){8-15}
 & Direction & Search & Compare & 1-Day & M-Day & \textbf{Mean} 
 & Frm. & Tool. & Cov. & Rel. & Acc. & Dep. & Cla. & \textbf{Mean (Val.\,\%)} & \\
\midrule
GPT-5.2  & 20.4 & 57.1 & 16.4 & 31.7 & 21.4 & 29.4 & 76.5 & 67.1 & 59.4 & 70.6 & 73.5 & 77.6 & 50.0  & 67.8 (85.0) &  48.6\\
Grok-4 & 17.0 & 21.3 & 9.7 & 24.7 &  11.3 & 16.8 & 33.7 & 36.8 & 43.4 & 36.8 & 39.2 & 36.1 & 17.5 & 34.8 (83.0) & 25.8 \\
Gemini-2.5-Pro & 8.6 & 12.5 & 7.4 & 11.9 & 12.4  & 10.6 & 15.8 & 19.0 & 17.9 & 32.6 & 28.3 & 45.7 & 38.6 & 28.3 (92.0) & 19.5\\
Claude-4.5-Sonnet & 30.0 & 29.6 & 25.3 & 50.4 &  42.1 & 35.5 & 75.8 & 69.7 & 80.8 & \textbf{83.3} & \textbf{79.3} & \textbf{86.9} & \textbf{66.7} & \textbf{77.5 (99.0)}  & 56.5\\
\midrule
\multicolumn{15}{c}{\textit{\textbf{Cold start + RL}}} \\
\midrule
SFT  & 10.6 & 29.7 & 14.1 & 20.4 & 7.1  & 16.4 & 14.1 & 20.3 & 23.4 & 14.1 & 15.6 & 15.6 & 14.1 & 16.7 (32.0) & 16.6 \\
GRPO & 11.0 & 26.3 & 14.3 & 21.9 & 8.6  & 16.4 & 20.6 & 35.3 & 35.3 & 23.5 & 23.5 & 26.5 & 11.8 & 25.2 (17.0)  & 20.8\\
GSPO & 10.0 & 30.6 & 13.1 & 21.1 & 11.4 & 17.2 & 23.8 & 33.3 & 40.5 & 16.7 & 21.4 & 31.0 & 9.5  & 25.2 (21.0)  & 21.2\\
Reinforce++ & 11.2 & 28.8 & 14.5 & 20.3 & 9.2 & 16.8 & 21.6 & 31.5 & 33.5 & 20.5 & 21.6 & 22.8 & 12.7  & 23.5 (23.0) & 20.2\\
Writing-zero & 17.3 & 43.2 & 20.8 & 33.1 & 14.5 & 25.8 & 39.4 & 49.4 & 46.0 & 37.2 & 40.4 & 35.1 & 33.0  & 40.1 (88.0)  & 33.0\\
Pref-GRPO & 25.7 & 50.4 & 27.5 & 41.2 & 16.4 & 32.2 & 52.6 & 56.3 & 50.5 & 44.3 & 44.3 & 46.7 & 36.5  & 47.3 (96.0) & 39.8 \\
\midrule
\multicolumn{15}{c}{\textit{\textbf{Direct RL}}} \\
\midrule
ArenaRL & 30.6 & 47.0 & 20.9 & 42.6 & 38.4 & 35.9 & 64.6 & 74.2 & 67.7 & 45.5 & 47.0 & 42.4 & 41.4  & 54.7 (99.0) & 45.3\\
\rowcolor{blue!12} \textbf{ArenaFlow} 
     & \textbf{43.3} & \textbf{72.9} & \textbf{45.1} & \textbf{64.1} & \textbf{51.6} & \textbf{55.4}
     & \textbf{76.8} & \textbf{87.9} & \textbf{82.3} & 58.1 & 65.2 & 59.1 & 54.0 & 69.1 (99.0)  & \textbf{62.3} \\
\bottomrule
\end{tabular}
}
\end{table*}

As shown in Table~\ref{tab:main}, ArenaFlow achieves the best overall performance across Open-Travel and Open-DeepResearch. 
On Open-Travel, ArenaFlow substantially outperforms both closed-source models and RL baselines, improving the mean score over ArenaRL from 35.9 to 55.4. 
On Open-DeepResearch, ArenaFlow achieves the strongest performance among RL baselines, improving over ArenaRL from 54.7 to 69.1. 
These consistent gains indicate that hierarchical credit propagation provides more effective supervision than trajectory-level preference optimization alone.
Notably, ArenaFlow also surpasses cold-start-then-RL methods by a large margin, improving over Pref-GRPO by 23.2 points on Open-Travel and 21.8 points on Open-DeepResearch. 
This demonstrates that ArenaFlow can achieve strong open-ended agent optimization without relying on expensive supervised cold-start data. 
Overall, these results support our motivation that rich comparative feedback should be propagated to pivotal reasoning steps and reusable skills, rather than being compressed into a single trajectory-level reward.

Table~\ref{tab:deepresearch_bench} further evaluates the model trained on Open-DeepResearch on DeepResearch Bench. 
ArenaFlow obtains the best overall score of 43.2, outperforming ArenaRL by 5.6 and Pref-GRPO by 9.1. 
It also performs competitively against proprietary deep-search systems such as Grok Deeper Search and Perplexity Deep Research. 
This transfer result suggests that ArenaFlow learns reusable research strategies rather than benchmark-specific behaviors. 
By maintaining high-utility skills and retrieving them as exploration priors, ArenaFlow improves the agent's ability to generalize to more challenging research-oriented tasks.

\subsection{Ablation Study}

\begin{table}[t]
\centering
\caption{Transfer evaluation on DeepResearch Bench.}
\label{tab:deepresearch_bench}
\resizebox{0.6\linewidth}{!}{
\setlength{\tabcolsep}{3pt}
\begin{tabular}{l ccccc}
\toprule
\multirow{2}{*}{\textbf{Method}} & \multicolumn{5}{c}{\textbf{DeepResearch Bench}} \\
\cmidrule(lr){2-6}
 & Overall & Comp. & Depth & Inst. & Read. \\
\midrule
Gemini-2.5-Pro & 35.1 & 34.1 & 29.8 & 41.7 & 37.2 \\
Grok Deeper Search & 40.2 & 38.0 & 35.4 & 46.3 & 44.1 \\
Perplexity Deep Research & 42.3 & 40.7 & 39.4 & 46.4 & 44.3 \\
\midrule
GRPO & 28.6 & 25.1 & 26.8 & 33.5 & 33.7 \\
Pref-GRPO & 34.1 & 31.6 & 28.4 & 40.7 & 38.3 \\
\midrule
ArenaRL & 37.6 & 35.8 & 36.0 & 40.5 & 40.0 \\
\rowcolor{blue!12}
\textbf{ArenaFlow} 
     & \textbf{43.2} & \textbf{41.7} & \textbf{40.8} & \textbf{46.7} & \textbf{45.8} \\
\bottomrule
\end{tabular}
}
\end{table}

\begin{table}[t]
\centering
\caption{Ablation study of different components of ArenaFlow on Open-Travel benchmark.}
\label{tab:ablation}
\resizebox{0.6\linewidth}{!}{
\setlength{\tabcolsep}{3pt}
\begin{tabular}{l ccccc c}
\toprule
\multirow{2}{*}{\textbf{Method}} & \multicolumn{5}{c}{\textbf{Open-Travel}} & \multirow{2}{*}{\textbf{Mean}} \\
\cmidrule(lr){2-6}
 & Direction & Search & Compare & 1-Day & M-Day & \\
\midrule
w/o Step Credit & 36.8 & 65.5 & 38.7 & 53.2 & 48.8 & 48.6 \\
w/o Skill Credit & 40.5 & 68.6 & 35.1 & 56.9 & 52.2 & 50.7 \\
w/o Skill Memory & 31.8 & 58.4 & 29.6 & 44.1 & 42.3 & 41.2 \\
w/o Depth Weight & 38.2 & 68.6 & 40.4 & 60.2 & 48.2 & 51.1 \\
\midrule
\rowcolor{blue!12}
\textbf{ArenaFlow} 
     & \textbf{43.3} & \textbf{72.9} & \textbf{45.1} & \textbf{64.1} & \textbf{51.6} & \textbf{55.4} \\
\bottomrule
\end{tabular}
}
\end{table}

We conduct ablation studies on Open-Travel to quantify the contribution of each component in ArenaFlow. 
As shown in Table~\ref{tab:ablation}, \textit{w/o Step Credit} removes step credit propagation and optimizes the policy only with trajectory-level advantages. 
\textit{w/o Skill Credit} disables usage-attributed utility estimation, so extracted skills are not updated or pruned. 
\textit{w/o Skill Memory} removes the entire skill memory module. 
\textit{w/o Depth Weight} replaces the tournament-depth weighting with a uniform weight.
Removing step-level credit reduces the mean score from 55.4 to 48.6, showing that trajectory-level advantages alone suffer from credit dilution in long-horizon planning tasks. 
Disabling skill-level credit lowers the mean score to 50.7, suggesting that usage-attributed utility estimation is necessary to suppress stale or ineffective skills. 
The largest degradation occurs when the skill memory is entirely removed, which highlights the importance of reusable skills as exploration priors.
In addition, \textit{w/o Depth Weight} further validates that pivotal steps identified in deeper tournament rounds provide more reliable credit signals. 
Overall, the ablation results show that ArenaFlow benefits from the synergy between step-level and skill-level credit propagation. 
Step-level credit provides targeted supervision for decisive reasoning behaviors, while skill-level credit consolidates successful behaviors into reusable priors for future exploration.

\subsection{Further Analysis}

\begin{table}[t]
\centering
\caption{Cross-judge robustness of ArenaFlow on Open-Travel with different reward models.}
\label{tab:reward_model}
\resizebox{0.6\linewidth}{!}{
\setlength{\tabcolsep}{3pt}
\begin{tabular}{l ccccc c}
\toprule
\multirow{2}{*}{\textbf{Model}} & \multicolumn{5}{c}{\textbf{Open-Travel}} & \multirow{2}{*}{\textbf{Mean}} \\
\cmidrule(lr){2-6}
 & Direction & Search & Compare & 1-Day & M-Day & \\
\midrule
Qwen3-Max & 43.3 & 72.9 & 45.1 & 64.1 & 51.6 & 55.4 \\
GPT-5 & \textbf{46.8} & 67.5 & 48.1 & 65.6 & 56.2 & 56.8 \\
Claude-4.5-Sonnet & 46.2 & \textbf{75.4} & \textbf{53.6} & \textbf{65.6} & \textbf{57.2} & \textbf{59.6} \\
\bottomrule
\end{tabular}
}
\end{table}


\paragraph{Reward Model Robustness.} We evaluate the robustness of ArenaFlow by replacing the reward model used in RL training. 
As shown in Table~\ref{tab:reward_model}, ArenaFlow consistently performs well with Qwen3-Max, GPT-5, and Claude-4.5-Sonnet as reward models. 
This indicates that ArenaFlow's improvement is not tied to a specific judge family, but stems from its general mechanism of converting comparative feedback into step-level and skill-level credit signals.
Moreover, stronger judges generally lead to higher performance, suggesting that higher-quality reflective feedback can more accurately identify pivotal steps and useful skills.

\begin{wrapfigure}{r}{0.45\textwidth}
\vspace{-15pt}
    \centering
    \includegraphics[width=\linewidth]{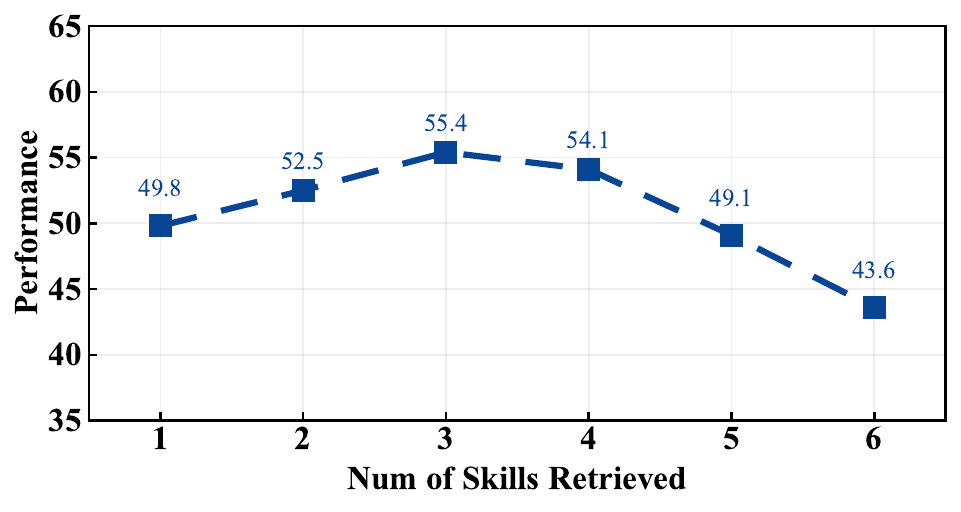}
    \caption{
    Effect of the retrieved skill count.}  
    \label{fig:skill_num}
    \vspace{-40pt}
\end{wrapfigure}

\paragraph{Effect of the Number of Retrieved Skills.}
Figure~\ref{fig:skill_num} studies the effect of the number of retrieved skills on Open-Travel. 
Performance improves from 49.8 to 55.4 as the number of retrieved skills increases from 1 to 3, and then steadily declines to 43.6 when 6 skills are injected. 
This reveals a trade-off in skill retrieval: too few skills provide insufficient strategic guidance, whereas too many skills introduce redundancy, distract the policy model, and increase prompt interference. 
The best performance at $k=3$ suggests that a compact set of high-utility skills is sufficient to provide effective exploration guidance.


\begin{wrapfigure}{r}{0.45\textwidth}
\vspace{-12pt}
    \centering
    \includegraphics[width=\linewidth]{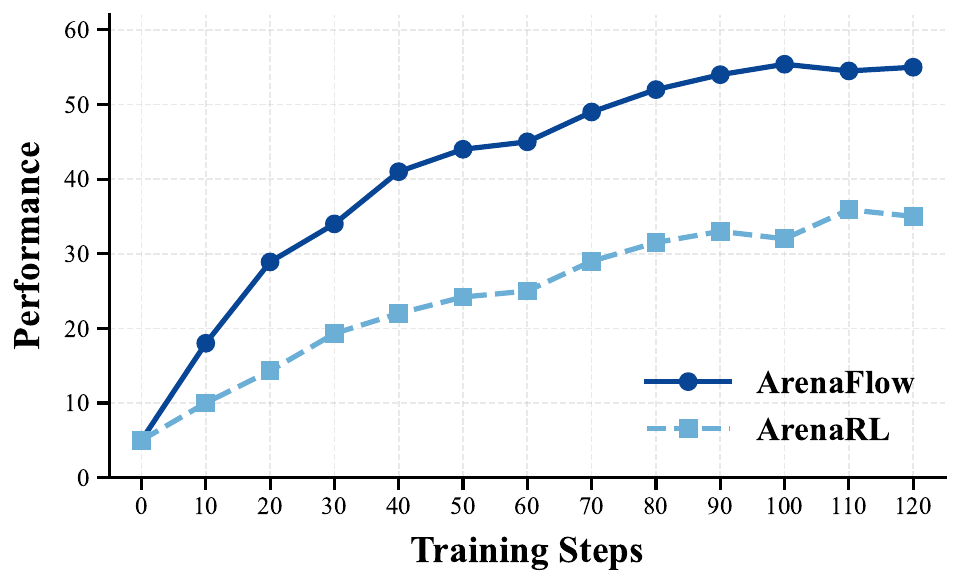}
    \caption{
    Training curves of ArenaFlow and ArenaRL on Open-Travel.}  
    \label{fig:training_curve}
    \vspace{-20pt}
\end{wrapfigure}

\paragraph{Training Efficiency.}
Figure~\ref{fig:training_curve} compares the training curves of ArenaFlow and ArenaRL on Open-Travel benchmark. 
ArenaFlow improves substantially faster, reaching 28.9 within 20 training steps, while ArenaRL only reaches 14.3. 
ArenaFlow also converges to a much higher final performance, around 55.4 compared with 35.9 for ArenaRL. 
This indicates that hierarchical credit propagation provides more effective learning signals than trajectory-level preference optimization alone. 
By assigning credit to pivotal reasoning steps and accumulating reusable skills as exploration priors, ArenaFlow improves both sample efficiency and final task performance.

More detailed analyses, including computational cost, structured reflective evaluation reliability, skill utility weighting, and skill memory pruning, are provided in Appendix~\ref{appendix:further_analysis}.
\section{Conclusion}

In this paper, we introduced ArenaFlow, a hierarchical credit propagation framework for open-ended agent RL. 
Instead of collapsing comparative feedback into a single trajectory-level reward, ArenaFlow propagates preference signals to pivotal reasoning steps for targeted policy optimization and to reusable skills for long-term exploration guidance. 
Extensive experiments on open-ended agent benchmarks demonstrate that ArenaFlow provides more effective supervision and enhances the agent's planning capabilities.

\clearpage
{
\bibliography{ArenaRL}
\bibliographystyle{colm2024_conference}
}

\newpage
\appendix

\section{More Details about Experiments}

\subsection{Benchmark Details}
\label{appendix:benchmark}
\paragraph{Open-Travel.}
Open-Travel evaluates tool-augmented agents on personalized itinerary planning. 
It covers five subtasks: multi-waypoint routing (\textit{Direction}), single-day itinerary planning (\textit{1-Day}), transportation comparison (\textit{Compare}), nearby point-of-interest discovery (\textit{Search}), and multi-day itinerary planning (\textit{M-Day}). 
The benchmark emphasizes multi-constraint reasoning, tool coordination, and alignment with user preferences. 
It provides six travel-related tools, including POI search, around search, navigation, universal search, flight search, and train-ticket search.

\paragraph{Open-DeepResearch.}
Open-DeepResearch evaluates agents on multi-turn web search and information synthesis for open-ended research tasks. 
It covers three representative scenarios: technical documentation drafting, research-plan formulation, and accessible explanation of complex concepts. 
Agents are allowed to interact with web search tools based on the Google API.

\paragraph{DeepResearch Bench.}
DeepResearch Bench contains 100 PhD-level research tasks designed by domain experts across 22 fields, including science and technology, finance and business, software engineering, and art and design. 
It evaluates whether agents can conduct deep information seeking and generate high-quality research reports. 
We use DeepResearch Bench as a transfer benchmark to evaluate models trained on Open-DeepResearch.

\paragraph{Evaluation Protocol.}
For Open-Travel and Open-DeepResearch, we adopt an LLM-as-a-Judge protocol with Qwen3-Max~\cite{yang2025qwen3} and Claude-4-Sonnet~\cite{claude} as dual judges. 
Each model trajectory is pairwise compared with the benchmark baseline trajectory. 
The judges evaluate both the reasoning trajectory and the final answer under task-specific rubrics. 
We compute win rates over non-tied comparisons and average the scores from the two judges as the final metric.

For Open-Travel, we report the performance of each subtask and their mean score. 
For Open-DeepResearch, we report the valid generation rate, denoted as Val.~(\%). 
Under valid generation conditions, we compute rubric-specific win rates and report their average as the mean score.  
For DeepResearch Bench, we adopt Gemini-2.5-Pro as the evaluation model and follow the official RACE metric. 
RACE evaluates the generated report against a reference report along four dimensions: Comprehensiveness (Comp.), Insight/Depth (Insight), Instruction-Following (Inst.), and Readability (Read.).

\subsection{Compared Baselines.}
\label{appendix:baseline}

\paragraph{Baseline Methods.}
We compare ArenaFlow with representative RL baselines for open-ended agent optimization. 
\textbf{GRPO}~\cite{grpo} estimates relative advantages within a sampled response group, avoiding the need for an explicit value model. 
\textbf{GSPO}~\cite{gspo} extends group-based policy optimization by performing sequence-level clipping, rewarding, and optimization. 
\textbf{Reinforce++}~\cite{reinforce++} improves reinforcement learning stability through global reward normalization and KL regularization. 
\textbf{Writing-Zero}~\cite{jia2025writing} assigns binary positive or negative advantages by comparing responses against random references. 
\textbf{Pref-GRPO}~\cite{wang2025pref} derives preference rewards from win rates obtained through exhaustive pairwise comparisons. 
\textbf{ArenaRL}~\cite{arenarl} formulates open-ended agent RL as an intra-group tournament ranking problem and performs pairwise evaluation over both reasoning trajectories and final answers.

\paragraph{Implementation.}
GRPO, GSPO, and Reinforce++ are implemented as pointwise reward baselines. 
They follow the standard LLM-as-a-Judge setting, where scalar rewards are obtained by scoring final answers. 
For fairness, these methods use the same judge models and evaluation rubrics as ArenaFlow. 
Writing-Zero and Pref-GRPO are implemented as pairwise reward baselines, but their comparisons are conducted only over final answers. 
ArenaRL also adopts pairwise reward, but evaluates both reasoning trajectories and final answers during tournament ranking.

\paragraph{Training Data.}
All trainable baselines use Qwen3-8B~\cite{yang2025qwen3} as the backbone. 
For Open-Travel and Open-DeepResearch, models in the cold-start setting are first fine-tuned on the corresponding SFT datasets to acquire basic tool-use and planning abilities. 
During RL, all baselines are trained on the corresponding RL splits of Open-Travel and Open-DeepResearch.

\section{More Results}
\label{appendix:further_analysis}

\begin{table}[t]
\centering
\caption{
Computational cost comparison on Open-Travel benchmark.
}
\label{tab:cost_analysis}
\resizebox{0.7\linewidth}{!}{
\large
\setlength{\tabcolsep}{4.5pt}
\begin{tabular}{lcccc}
\toprule
\textbf{Method} & \textbf{Training Time} & \textbf{Tokens (M)} & \textbf{LLM Call Count} & \textbf{Performance} \\
\midrule
GRPO       & 57 min      & 0.29 & 128 & 16.4\% \\
Pref-GRPO  & 2 h 02 min  & 3.87 & 960 & 32.2\% \\
ArenaRL    & 1 h 16 min  & 1.52 & 240 & 35.9\% \\
ArenaFlow  & 1 h 22 min  & 1.97 & 248 & 55.4\% \\
\bottomrule
\end{tabular}
}
\end{table}

\paragraph{Computational Cost Analysis.}
Table~\ref{tab:cost_analysis} compares the computational cost of different RL methods on Open-Travel under the same setting with group size $N=16$ and group number $G=8$. 
ArenaFlow incurs only moderate additional overhead compared with ArenaRL, increasing token usage from 1.52M to 1.97M and LLM calls from 240 to 248. 
Despite this small increase, ArenaFlow improves performance from 35.9\% to 55.4\%. 
Compared with Pref-GRPO, ArenaFlow requires substantially fewer LLM calls and tokens while achieving much higher performance. 
These results indicate that ArenaFlow extracts richer step-level and skill-level supervision from tournament comparisons, yielding a favorable trade-off between efficiency and performance.

\paragraph{Reliability of Structured Reflective Evaluation.}
Since ArenaFlow relies on structured reflective evaluation to propagate credit to pivotal reasoning steps and attributed skills, we further examine the reliability of these two feedback signals.
Specifically, we randomly sample 200 pairwise comparisons from Open-Travel and Open-DeepResearch, and hire three human annotation experts to label pivotal success steps and skill usage attribution.
For pivotal step identification, annotators are instructed to mark a step as pivotal only if it directly contributes to the winning trajectory, such as resolving a key constraint, correcting an erroneous plan, or retrieving decisive evidence.
For skill usage attribution, annotators determine whether a retrieved skill is substantively reflected in these pivotal steps, rather than merely being semantically related to the query or appearing in the prompt.
The majority-voted annotations are used as reference labels.

As shown in Table~\ref{tab:judge_feedback_reliability}, the judge-generated feedback aligns well with human annotations, achieving F1 scores of 79.5 and 76.1 for pivotal step identification and skill usage attribution, respectively.
The average F1 score of 77.8 indicates that structured reflective feedback can provide reliable supervision for hierarchical credit propagation.
This further suggests that ArenaFlow's improvements are not merely driven by noisy post-hoc rationales, but by structured feedback signals that are broadly consistent with human judgment.

\begin{wrapfigure}{r}{0.45\textwidth}
\vspace{-15pt}
    \centering
    \includegraphics[width=\linewidth]{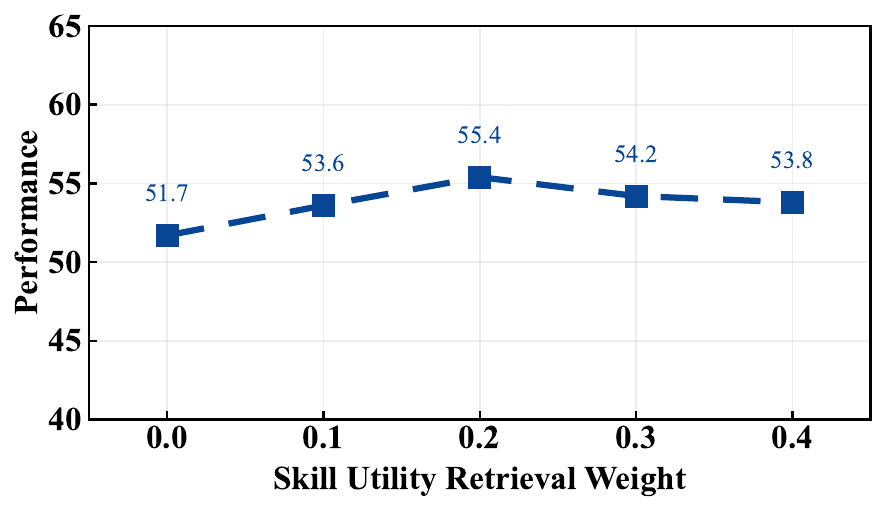}
    \caption{
    Effect of the skill utility weight in retrieval reranking on Open-Travel benchmark.}  
    \label{fig:utility_weight}
    \vspace{-30pt}
\end{wrapfigure}

\paragraph{Effect of Skill Utility Weight.}
We further study the effect of the skill utility weight in the retrieval reranking score on Open-Travel benchmark. 
As shown in Figure~\ref{fig:utility_weight}, performance improves as the utility weight increases from $0.0$ to $0.2$, with the mean score rising from 51.7 to 55.4. 
This indicates that incorporating historical utility helps the retriever select skills that are not only semantically relevant but also empirically effective in previous successful trajectories. 
However, further increasing the utility weight leads to slightly lower performance, suggesting that overemphasizing historical utility may weaken query-specific relevance and cause the agent to retrieve generally useful but less task-matched skills. 
The best performance at $0.2$ shows that effective skill retrieval requires balancing semantic relevance and usage-attributed utility.

\begin{table}[t]
\centering
\footnotesize
\setlength{\tabcolsep}{6pt}
\caption{Reliability analysis of structured reflective feedback.}
\label{tab:judge_feedback_reliability}
\begin{tabular}{lccc}
\toprule
\textbf{Feedback Type} & \textbf{Precision} & \textbf{Recall} & \textbf{F1} \\
\midrule
Pivotal Step Identification & 82.4 & 76.8 & 79.5 \\
Skill Usage Attribution     & 78.9 & 73.5 & 76.1 \\
\midrule
Average                     & 80.7 & 75.2 & 77.8 \\
\bottomrule
\end{tabular}
\end{table}


\begin{wrapfigure}{r}{0.45\textwidth}
\vspace{-12pt}
    \centering
    \includegraphics[width=\linewidth]{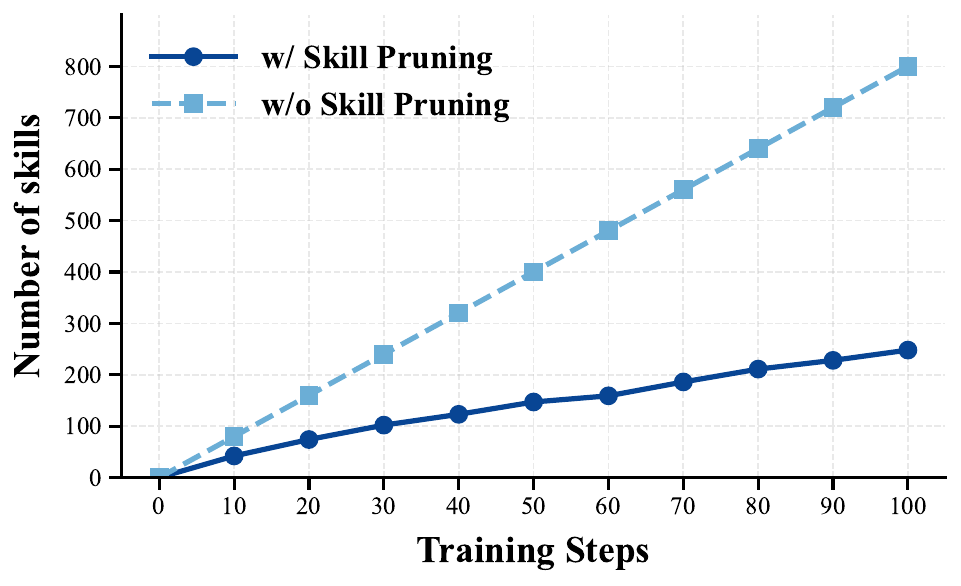}
    \caption{
    Evolution of skill memory size with and without utility-aware skill pruning during RL training.}  
    \label{fig:skill_pruning}
    \vspace{-20pt}
\end{wrapfigure}

\paragraph{Effect of Skill Pruning.}
Figure~\ref{fig:skill_pruning} compares the evolution of skill memory size with and without utility-aware skill pruning. 
Without pruning, the number of stored skills grows nearly linearly and reaches 800 after 100 training steps. 
In contrast, utility-aware pruning keeps the memory substantially more compact, retaining about 248 skills at the end of training. 
This demonstrates that pruning is important for preventing uncontrolled memory expansion and removing redundant or low-utility skills. 
By maintaining a compact and higher-quality skill memory, ArenaFlow reduces retrieval noise and preserves more reliable exploration priors for subsequent rollouts.


\begin{figure*}[ht]
    \centering
    \includegraphics[width=1\textwidth]{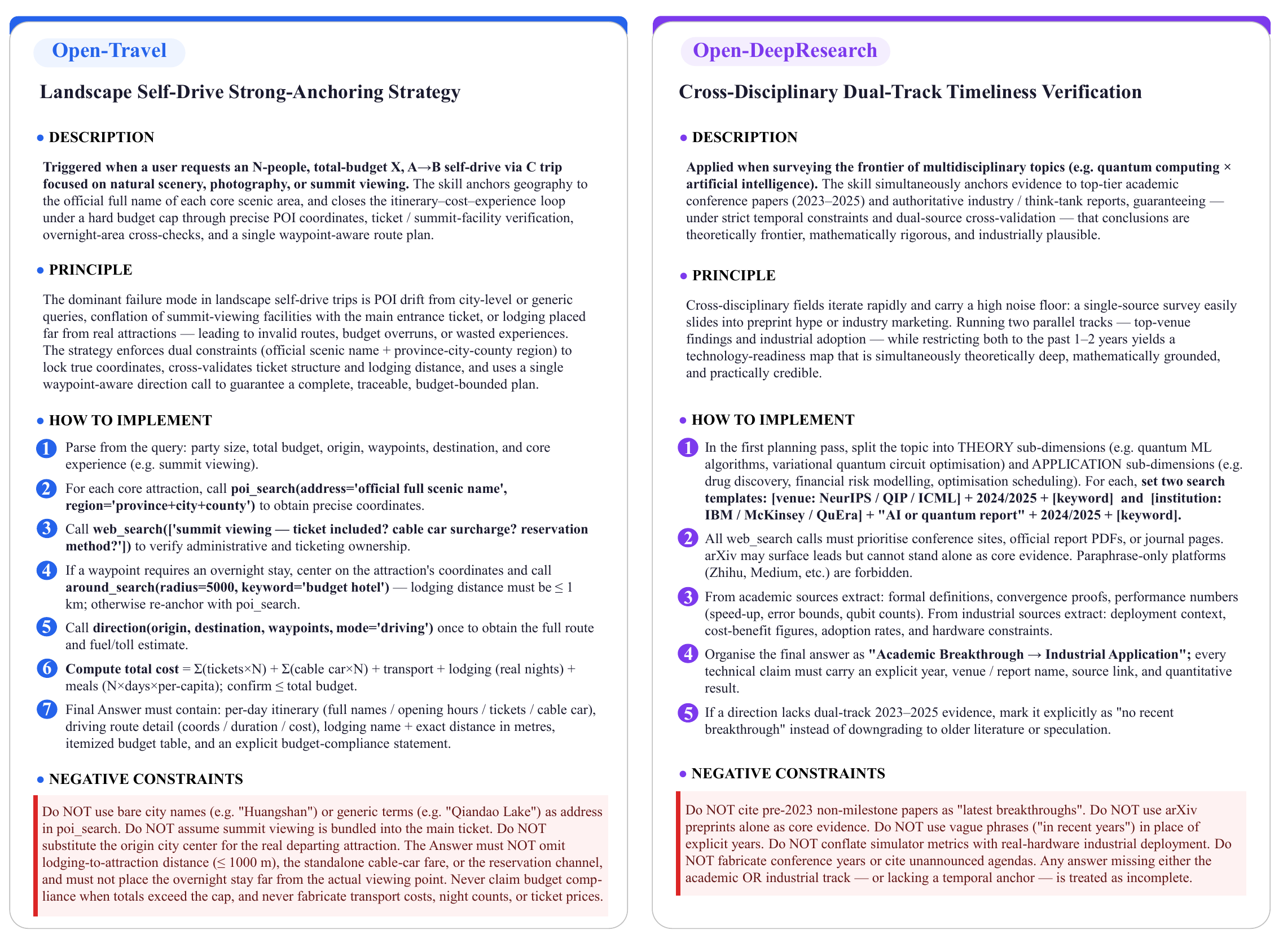}
    \caption{Examples of reusable skills learned by ArenaFlow on Open-Travel and Open-DeepResearch. }
    \label{fig:skill_example}
\end{figure*}

\section{Visualization of Skills}
Figure~\ref{fig:skill_example} presents representative skills learned by ArenaFlow from Open-Travel and Open-DeepResearch. 
The Open-Travel skill converts landscape-oriented self-drive itinerary planning into a structured tool-use strategy, covering POI grounding, ticket verification, lodging search, route planning, and budget checking. 
The Open-DeepResearch skill captures a reusable evidence-verification strategy that distinguishes academic and industrial sources, prioritizes recent authoritative evidence, and organizes responses from theoretical foundations to practical applications. 
Both skills include explicit failure-avoidance constraints to prevent common errors such as ambiguous POI grounding, fabricated costs, outdated citations, and unsupported claims. 
These examples show that ArenaFlow can distill successful trajectories into compact and reusable priors that combine high-level principles with concrete execution guidelines.

\section{Case Study}
\label{appendix:case_study}
Figure~\ref{fig:case_smart_anchor_bolt} presents a representative Open-DeepResearch case to illustrate how ArenaFlow solves open-ended research tasks.
The case highlights ArenaFlow's long-horizon planning ability.
The agent decomposes the user query into a sequence of goal-directed stages, including problem analysis, evidence collection, information filtering, and final synthesis. 
During this process, ArenaFlow dynamically refines its search focus according to the collected evidence, ensuring that each tool call contributes to a specific aspect of the research objective. 
The final response integrates evidence from multiple sources into a coherent and well-structured answer. 
This demonstrates that ArenaFlow can maintain task coherence while coordinating search, reasoning, and synthesis in open-ended research scenarios.

\section{Prompts}
\label{appendix:prompt}
In this section, we provide the prompt templates used in different stages of ArenaFlow, with the Chinese prompts translated into English.

\subsection{System Prompts}
Figure~\ref{fig:system prompt open-travel} presents the system prompt used by the agent on Open-Travel, and Figure~\ref{fig:system prompt open-deepresearch} shows that for Open-DeepResearch.

\subsection{Open-DeepResearch Judge Prompt}
Figure~\ref{fig:prompt-judge-open-travel} shows the judge prompt used for pairwise trajectory comparison on Open-Travel, and Figure~\ref{fig:judge-prompt-open-deepresearch} shows that for Open-DeepResearch.
The judge evaluates both the reasoning trajectory and the final answer, and outputs structured reflective feedback for ArenaFlow.

\subsection{Skill Merge Prompt}
Figure~\ref{fig:skill-merge-prompt} shows the prompt used to merge and refine skills extracted from the champion trajectory. 
The prompt consolidates redundant or fragmented skill descriptions into a compact and reusable skill entry for the global skill memory.


\clearpage
\onecolumn

\begin{tcolorbox}[breakable,title=Case Trajectory in Open-DeepResearch Benchmark Following ArenaFlow (translated into English), label=case:smart_anchor_bolt]
\textcolor{black}{\textbf{Query:} Conduct an in-depth market study on smart anchor bolts. I need to write a business plan and develop a smart anchor bolt product.}\\

\colorbox{gray!70}{%
    \parbox{\dimexpr\textwidth-2\fboxsep}{%
    \centering
        \textbf{Step 1}%
    }%
}
\textcolor{black}{\textbf{<think>}\\
The user request involves two core objectives: conducting market research for a business plan and developing a smart anchor bolt product. The task should be organized around technical specifications, application cases, and regulatory compliance, while covering technology, market, business model, policy, risks, and future trends. ...\\
I will first focus on identifying the technical architecture and core parameters of smart anchor bolts, including sensor types, data transmission protocols, material strength indicators, patent numbers, standard references, and quantitative performance data.\\
\textbf{</think>}}\\

\textcolor{purple1}{\textbf{<tool\_call>}\\
\{"name": "web\_search", "arguments": \{"query": ["smart anchor bolt core technical parameters 2025", "smart anchor bolt sensor types material strength standards", "smart anchor bolt patent number performance indicators"]\}\}\\
\textbf{</tool\_call>}}\\

\textcolor{black}{\textbf{<tool\_response>}\\
...\\
\textbf{</tool\_response>}}\\

\colorbox{gray!70}{%
    \parbox{\dimexpr\textwidth-2\fboxsep}{%
    \centering
        \textbf{Step 2}%
    }%
}
\textcolor{black}{\textbf{<think>}\\
The previous step has obtained key technical parameters and patent information for smart anchor bolts. The findings cover sensor types, material strength, deformation capacity, real-time displacement monitoring, and representative patents such as CN102877870A and 201010618198.1. ...\\
Next, I will move to application scenarios and customer cases, focusing on mining, tunnel engineering, and other infrastructure settings, and extracting concrete institutions, implementation contexts, and quantitative outcomes.\\
\textbf{</think>}}\\

\textcolor{purple1}{\textbf{<tool\_call>}\\
\{"name": "web\_search", "arguments": \{"query": ["smart anchor bolt application scenarios cases 2025", "smart anchor bolt mining case implementation effect", "smart anchor bolt tunnel engineering quantitative outcomes"]\}\}\\
\textbf{</tool\_call>}}\\

\textcolor{black}{\textbf{<tool\_response>}\\
...\\
\textbf{</tool\_response>}}\\

\colorbox{gray!70}{%
    \parbox{\dimexpr\textwidth-2\fboxsep}{%
    \centering
        \textbf{Step 3}%
    }%
}
\textcolor{black}{\textbf{<think>}\\
The previous step has collected representative application scenarios and case evidence for smart anchor bolts. The results include mining applications such as Jinqiao Coal Mine in Jining, Shandong Province, as well as tunnel-engineering cases involving intelligent support decision systems. ...\\
These findings provide evidence for typical use cases and measurable benefits. I will now proceed to policy, regulation, and industry-standard analysis, with emphasis on work safety, data security, cybersecurity, and anchor-bolt support specifications.\\
\textbf{</think>}}\\

\textcolor{purple1}{\textbf{<tool\_call>}\\
\{"name": "web\_search", "arguments": \{"query": ["smart anchor bolt policies and regulations 2025", "smart anchor bolt data security standards", "anchor-bolt support work-safety specifications"]\}\}\\
\textbf{</tool\_call>}}\\

\textcolor{black}{\textbf{<tool\_response>}\\
...\\
\textbf{</tool\_response>}}\\

\colorbox{gray!70}{%
    \parbox{\dimexpr\textwidth-2\fboxsep}{%
    \centering
        \textbf{Step 4}%
    }%
}
\textcolor{black}{\textbf{<think>}\\
The previous step has summarized relevant policies, regulations, and safety standards, including intelligent construction guidelines, MT/T 1104--2009, and underground mine support safety requirements. ...\\
Since the compliance dimension has been covered, I will now analyze the competitive landscape by identifying leading companies, product features, pricing strategies, distribution channels, and potential sources of differentiation.\\
\textbf{</think>}}\\

\textcolor{purple1}{\textbf{<tool\_call>}\\
\{"name": "web\_search", "arguments": \{"query": ["smart anchor bolt market competition 2025", "smart anchor bolt leading companies product analysis", "anchor-bolt support pricing strategy distribution model"]\}\}\\
\textbf{</tool\_call>}}\\

\textcolor{black}{\textbf{<tool\_response>}\\
...\\
\textbf{</tool\_response>}}\\

\colorbox{gray!70}{%
    \parbox{\dimexpr\textwidth-2\fboxsep}{%
    \centering
        \textbf{Step 5}%
    }%
}
\textcolor{black}{\textbf{<think>}\\
The previous step has identified the competitive landscape and representative companies in the smart anchor-bolt and bolting-equipment market, including Sandvik, Hunan Wuxin Tunnel Intelligent Equipment, Sichuan Lanhai Intelligent Equipment, and other major suppliers. ...\\
Next, I will construct the business model framework, focusing on pricing models, revenue channels, direct sales, partner networks, subscription-based services, and the broader cooperation ecosystem.\\
\textbf{</think>}}\\

\textcolor{purple1}{\textbf{<tool\_call>}\\
\{"name": "web\_search", "arguments": \{"query": ["smart anchor bolt business model 2025", "anchor-bolt support pricing strategy case", "smart anchor bolt channel model cooperation ecosystem"]\}\}\\
\textbf{</tool\_call>}}\\

\textcolor{black}{\textbf{<tool\_response>}\\
...\\
\textbf{</tool\_response>}}\\

\colorbox{gray!70}{%
    \parbox{\dimexpr\textwidth-2\fboxsep}{%
    \centering
        \textbf{Step 6}%
    }%
}
\textcolor{black}{\textbf{<think>}\\
The previous step has established a preliminary business model for smart anchor bolts, including cost-based pricing, value-based pricing, subscription services, direct sales, partner networks, and cooperation with sensor suppliers, software developers, and research institutions. ...\\
I will now evaluate the key risks of the proposed product and business plan, including technical risks, market risks, and policy risks, and then propose corresponding mitigation strategies.\\
\textbf{</think>}}\\

\textcolor{purple1}{\textbf{<tool\_call>}\\
\{"name": "web\_search", "arguments": \{"query": ["smart anchor bolt risk assessment 2025", "anchor-bolt support risk-control measures", "smart anchor bolt policy risk mitigation strategy"]\}\}\\
\textbf{</tool\_call>}}\\

\textcolor{black}{\textbf{<tool\_response>}\\
...\\
\textbf{</tool\_response>}}\\

\colorbox{gray!70}{%
    \parbox{\dimexpr\textwidth-2\fboxsep}{%
    \centering
        \textbf{Step 7}%
    }%
}
\textcolor{black}{\textbf{<think>}\\
The previous step has identified the main technical, market, and policy risks of smart anchor bolts, together with mitigation strategies such as redundant sensor design, encrypted data transmission, differentiated market positioning, long-term service agreements, and dynamic compliance tracking. ...\\
All planned research dimensions have now been completed. I will integrate the collected evidence into a structured business plan and R\&D proposal.\\
\textbf{</think>}}\\

\textcolor{purple1}{\textbf{<tool\_call>}\\
\{"name": "web\_search", "arguments": \{"query": ["six-dimensional report template for smart anchor bolts"]\}\}\\
\textbf{</tool\_call>}}\\

\textcolor{black}{\textbf{<tool\_response>}\\
...\\
\textbf{</tool\_response>}}\\

\colorbox{gray!70}{%
    \parbox{\dimexpr\textwidth-2\fboxsep}{%
    \centering
        \textbf{Step 8}%
    }%
}
\textcolor{black}{\textbf{<answer>}\\
\textbf{Business Plan and R\&D Proposal for Smart Anchor Bolts}\\
\textit{Based on a six-dimensional, three-anchor framework}\\
\textbf{1. Technical Architecture and Core Parameters}\\
\textbf{1.1 Technical Anchoring.}
The proposed smart anchor bolt integrates fiber Bragg grating sensors and displacement sensors to enable real-time monitoring of anchor-bolt displacement, stress, and temperature. Representative patents include CN102877870A and 201010618198.1. In terms of material strength, the product adopts a combination of fiber-reinforced plastic and high-strength steel, with the anchoring-body strength reaching at least 70\% of the design strength in accordance with GB 50086. Data transmission can rely on 5G or LoRa wireless communication to support real-time data upload and cloud-based analysis.\\
Key performance parameters include a constant resistance of 150 kN, adaptability to surrounding-rock deformation of 0.3--1.0 m, displacement monitoring accuracy of $\pm$1 mm, stress monitoring error of no more than 5\%, an operating temperature range from $-20^{\circ}$C to $80^{\circ}$C, and an IP68 waterproof rating.\\
\textbf{1.2 Technical Advantages.}
The product adopts a multi-stage yielding design, in which shear cores fracture progressively to realize three-level early warning. It also integrates intelligent algorithms for anchor-bolt parameter optimization, such as response surface methodology combined with genetic algorithms, which can reduce support costs by 15\% and decrease displacement by 22\%.\\
\textbf{2. Application Scenarios and Case Validation}\\
\textbf{2.1 Scenario Anchoring.}
Core application scenarios include deep hard-rock roadways and rockburst-prone areas in mining, mountain highway tunnels and metro shield-tunneling sections in tunnel engineering, as well as slope support and underground utility tunnels.\\
\textbf{2.2 Case Validation.}
In Jinqiao Coal Mine, Jining, Shandong Province, an intelligent anchor-bolt propulsion robot and real-time monitoring system were deployed. The system improved excavation efficiency by 30\%, reduced the workforce by 80 people, and decreased the accident rate by 65\%. In a tunnel project in Sichuan, an automatic anchor-bolt spacing optimization system was applied with a spacing configuration of 0.8 m $\times$ 0.8 m, reducing support costs by 15\% and roof displacement by 22\%.\\
\textbf{3. Policy Compliance and Standard Conformity}\\
\textbf{3.1 Policy Anchoring.}
Relevant national and industry standards include MT/T 1104--2009, \textit{Technical Specification for Anchor-Bolt Support in Coal Roadways}, and GB 50086, \textit{Technical Code for Ground Anchorage Engineering}. In terms of data security, the system should comply with the Cybersecurity Law through end-to-end encrypted transmission and with the Data Security Law through localized data storage.\\
\textbf{3.2 Compliance Measures.}
The product should obtain mining product safety certification, such as the MA mark. It should also comply with standard testing requirements, including anchor-bolt pull-out tests and anchoring-body strength requirements. For example, the anchoring-body strength should reach at least 70\% of the design strength according to relevant technical standards.\\
\textbf{4. Market Competition and Business Model}\\
\textbf{4.1 Market Anchoring.}
Major global competitors include Sandvik, Hunan Wuxin Tunnel Intelligent Equipment, and Sichuan Lanhai Intelligent Equipment. Domestic products have two major advantages: first, localization can exceed 70\%, enabling prices approximately 30\% lower than imported products; second, domestic suppliers can provide customized designs for complex geological conditions.\\
\textbf{4.2 Business Model.}
The pricing strategy may combine product sales and subscription services. For example, the product can be sold at a benchmark price of RMB 1,200 per unit, while data monitoring services can be charged at RMB 500 per month per system. Channel strategies include direct sales to large mining groups, such as China Coal Energy, regional agents for small and medium-sized customers, and online B2B platforms such as Made-in-China and HC360.\\
\textbf{4.3 Cooperation Ecosystem.}
The company can cooperate with sensor suppliers and software developers to build integrated intelligent support solutions. It can also collaborate with universities and research institutes, such as Tsinghua University and China University of Mining and Technology, to jointly develop new materials and intelligent algorithms, and participate in national key projects related to intelligent mining.\\
\textbf{5. Risk Control and Mitigation Strategies}\\
\textbf{5.1 Risk Anchoring.}
Technical risks include sensor failure and data transmission delays. Market risks include intensified competition and fluctuations in customer demand. Policy risks include regulatory changes and upgrades to safety standards.\\
\textbf{5.2 Mitigation Strategies.}
Technical risks can be mitigated through redundant design, dual-sensor verification, regular maintenance, localized data storage, and cloud backup. Market risks can be mitigated through differentiated positioning in deep mines and complex geological scenarios, customized services, and subscription-based pricing to retain long-term customers. Policy risks can be addressed by establishing a policy research team to track regulatory updates and by proactively preparing for new standards such as intelligent-mine construction guidelines.\\
\textbf{6. Future Trends and Strategic Outlook}\\
\textbf{6.1 Technology Trends.}
Future development should focus on AI integration, such as using large models for fault prediction and self-optimization. Green design is also important, including the development of recyclable anchor bolts to reduce environmental impact.\\
\textbf{6.2 Market Expansion.}
International expansion can target mining markets in Southeast Asia and Africa. Scenario expansion can extend the product to marine engineering, aerospace engineering, and other high-reliability support environments.\\
\textbf{6.3 Strategic Objectives.}
By 2025, the company aims to complete mass production of the core product and reach a 15\% market share. By 2027, it aims to cover major global mining areas and achieve revenue above RMB 500 million. By 2030, it aims to become a global leader in the smart anchor-bolt market with a market share exceeding 30\%.\\
\textbf{Appendix.}
Representative patents include CN102877870A, 201010618198.1, and 202210748162.8. Relevant standards include MT/T 1104--2009, GB 50086, and JGJ/T 182. Case sources include Jinqiao Coal Mine in Jining, the Sichuan tunnel engineering case, and relevant industrial cooperation models.\\
\textit{Note: This report follows the six-dimensional, three-anchor framework. All data and cases are organized from authoritative sources to ensure technical feasibility, market validity, and policy compliance.}\\
\textbf{</answer>}}\\

\end{tcolorbox}
\captionof{figure}{Case trajectory of ArenaFlow on the Open-DeepResearch benchmark.}
\label{fig:case_smart_anchor_bolt}

\begin{figure*}[!ht]
\centering
\begin{tcolorbox}[promptbox, title=System Prompt --- Open-Travel]
\begin{PromptVerb}
Current time: {time}

# Travel Planning Specification
You are a travel-planning research assistant. Your goal is to answer the user's query by decomposing it into a sequence of clear and executable steps.

## Tool Use
You may call tools for at most {max_steps} rounds. So far, {step_idx} rounds have been used.

## Relevant Skills
The following skills may be useful for solving similar tasks: {Relevant skills}
\end{PromptVerb}
\end{tcolorbox}
\caption{System prompt for Open-Travel benchmark.}
\label{fig:system prompt open-travel}
\end{figure*}

\begin{figure*}[!ht]
\centering
\begin{tcolorbox}[promptbox, title=System Prompt --- Open-DeepResearch]
\begin{PromptVerb}
Current time: {time}

# Deep Research Specification
You are a professional deep-research agent. Your core principles are rigor and clear reasoning. For any complex research question, you must not rush to provide a final answer in a single step. Instead, you must strictly follow a sequential workflow:

Plan -> Execute Step by Step -> Synthesize Final Report.

## Stage 1: Research Planning
This stage is performed only in the first response. Upon receiving the task, you must first conduct problem decomposition. Do not start searching immediately. Instead, carefully understand the user's intent and decompose the complex question into 3--6 logically connected and searchable sub-questions.

The first response must strictly follow this format:

[Problem Analysis]
Briefly analyze the user's core need, key concepts, and potential information gaps.

[Global Research Plan]
1. Step 1: Describe the specific sub-question or search objective.
2. Step 2: Describe the specific sub-question or search objective.
3. Step 3: Describe the specific sub-question or search objective.
...
Add up to 6 steps when necessary.

[First Action]
The global plan has been established. Next, I will focus on Step 1 and extract precise keywords for the initial search.

## Stage 2: Step-by-Step Investigation
From the second round to the penultimate round, you must avoid attempting to solve all sub-questions at once. Based on the previous search results, dynamically decide whether to further investigate the current step or move to the next step.

Before each search round, you must strictly follow this reasoning format:

- Previous-step assessment: Evaluate whether the last search for Step n was useful, what key information was obtained, whether it is sufficient for the goal of Step n, and what is still missing if it is insufficient.
- Current focus: Decide whether to continue supplementing Step n or proceed to Step n+1, and clearly state the objective of the current round.
- Search decision: Design precise search queries for the current focus, and explain the keyword design rationale.

## Stage 3: Final Report Synthesis
You may stop searching only when all steps have obtained sufficient information, or when the current tool-call round {step_idx} is close to the maximum number of rounds {max_steps}. All assessments in this stage should be performed silently in your internal reasoning. The final output should contain only a complete and detailed research report, without exposing any intermediate reasoning process.

## Report Quality Requirements
- Comprehensive: cover all key dimensions of the question with sufficient detail.
- Logical: present in-depth analysis with clear reasoning.
- Well-structured: organize content with headings, lists, or tables.
- Traceable: cite sources for key facts or data.
- Accurate: use precise domain terminology and avoid conceptual confusion.

## Relevant Skills
The following skills may be useful for solving similar tasks: {Relevant skills}
\end{PromptVerb}
\end{tcolorbox}
\caption{System prompt for Open-DeepResearch benchmark.}
\label{fig:system prompt open-deepresearch}
\end{figure*}

\clearpage

\begin{tcolorbox}[promptbox, title=Judge Prompt --- Open-Travel]
\begin{PromptVerb}
You are a comprehensive judge for travel-planning LLM agents, with expertise in the travel domain, rigorous reasoning, and evaluation methodology. Given the same user query, you will compare LLM Agent A and Agent B in terms of their reasoning paths and final answers.

The reasoning path refers to the complete ordered steps and tool-call logs. The final answer refers to the final response shown to the user. You must conduct dimension-wise comparative diagnosis and quantitative evaluation.

After comparative reflection, you must provide objective scores and a winner. Then, identify the pivotal success step of the winner, and abstract it into a generalizable and reusable skill. Strictly follow the criteria, scoring rules, and JSON output format below.

## Input Format

<USER_QUERY>{original user query}</USER_QUERY>

<RETRIEVED_SKILLS>{candidate skills injected into the system prompt during trajectory generation, including skill IDs and detailed contents. Leave empty if none.}</RETRIEVED_SKILLS>

<PATH_A>{complete reasoning path of LLM Agent A, with each step numbered}</PATH_A>

<PATH_B>{complete reasoning path of LLM Agent B, with each step numbered}</PATH_B>

<ANSWER_A>{complete final answer of LLM Agent A}</ANSWER_A>

<ANSWER_B>{complete final answer of LLM Agent B}</ANSWER_B>

## Path Evaluation

### Dimensions
1. Breadth: whether the path comprehensively covers the user's needs without redundant or repetitive steps.
2. Relevance: whether each step is well aligned with the user's core intent.
3. Detail: whether the cited facts, data, time points, prices, reservation rules, and other details are sufficient, accurate, and useful.

### Scoring Rules
- For path evaluation, focus only on actual tool calls in the reasoning path. Do not evaluate the depth of analysis over the retrieved information.
- Each dimension is scored from 0 to 10, where 0 means completely missing and 10 means excellent.
- The overall path score, `overall_p`, is the average of the three dimensions.

## Answer Evaluation

### Dimensions
1. Relevance: whether the answer fully addresses all sub-requirements and constraints, and whether its ordering fits the travel scenario.
2. Feasibility: whether the plan is logically consistent, practically executable, and free of obvious conflicts.
3. Details: whether the answer provides rich and useful information such as schedules, ticket prices, travel times, and practical tips.
4. Clarity: whether the structure, formatting, and readability are clear and user-friendly.

### Scoring Rules
- Evaluate the final answer with reference to the retrieved knowledge shown in the corresponding reasoning path.
- Each dimension is scored from 0 to 10, where 0 means completely missing and 10 means excellent.
- The overall answer score, `overall_a`, is the average of the four dimensions.

## Overall Score and Winner

The combined score is computed as:combined_score = 0.6 * overall_p + 0.4 * overall_a

Round the combined score to one decimal place. If the two combined scores are equal, the result is `Tie`.

## Output Format

Strictly output the following JSON object and do not add any extra content:

{
  "analysis": {
    "path_comparison": "<60-150 words. Provide a comparative diagnosis of the two paths. Explain the winner's key path advantages over the loser and the loser's weaknesses. Specify where the loser deviated from optimal planning, such as ineffective loops or parameter errors, and how the winner avoided them.>",
    "answer_comparison": "<60-150 words. Compare the final open-domain answers and explain the winner's concrete advantages in factual coherence, intent alignment, and information coverage.>"
  },
  "path_scores": {
    "Agent_A": {"breadth": <0-10>, "relevance": <0-10>, "detail": <0-10>, "overall_p": <0-10>},
    "Agent_B": {"breadth": <0-10>, "relevance": <0-10>, "detail": <0-10>, "overall_p": <0-10>}
  },
  "answer_scores": {
    "Agent_A": {"relevance": <0-10>, "feasibility": <0-10>, "details": <0-10>, "clarity": <0-10>, "overall_a": <0-10>},
    "Agent_B": {"relevance": <0-10>, "feasibility": <0-10>, "details": <0-10>, "clarity": <0-10>, "overall_a": <0-10>}
  },
  "combined_scores": {"Agent_A": <0-10>,"Agent_B": <0-10>},
  "winner": "<Agent_A | Agent_B | Tie>",
  "pivotal_success_step": {
    "trajectory": "<A | B>",
    "step_index": "<a list of integers or None, e.g., [3], [3,4], or None>",
    "referenced_skills": "<for each pivotal step, specify the referenced skill IDs from RETRIEVED_SKILLS that it references, e.g., [\"skill 0\"], [\"skill 0\", \"skill 1\"], or None. If a pivotal step relies only on native capability and no retrieved skill, use None at the corresponding position. If there is no pivotal step, output None.>",
    "reason": "<one concise sentence explaining what key action this step performed, why it decided the outcome, and why the referenced skill was relevant.>"
  },
  "skill": {
    "name": "<5-15 words, a concise tactical skill name>",
    "description": "<the triggering conditions, environment state, or semantic intent of the user query for applying this skill>",
    "principle": "<the core principle explaining why this strategy is logically superior>",
    "how_to_implement": "<a concrete execution guide. It may include tool-call formats, parameters, tool sequencing, and the final answer organization pattern, formulated as a standard operating procedure.>",
    "negative_constraints": "<pitfalls to avoid based on the loser's errors, covering both path-level and answer-level mistakes when applicable.>"
  }
}

## Important Requirements

- Think independently for each dimension before assigning scores to ensure fairness and objectivity.
- During scoring, completely ignore the content of <RETRIEVED_SKILLS>. Scores must be based only on the paths and final answers of Agent A and Agent B.
- All comments must be specific and traceable to the provided text. Do not introduce external information.
- The `step_index` field in `pivotal_success_step` must be a list, such as [5] or [5,6,7]. If no pivotal step exists, use None.
- The extracted skill must capture the most valuable generalizable experience, not merely summarize this instance.
- The `how_to_implement` field must include concrete tool-use rules and the final response organization paradigm.
- Focus on comparison: do not summarize A and B independently. Explain why one is better than the other in the same context, with concrete reference to step indices.
- Strictly follow the JSON template so that downstream programs can parse the output.

## Tool Explanations

- The `poi_search` tool searches for geospatial information about points of interest within a specified city.
- The `around_search` tool searches for places within a circular area by specifying a center point and radius.
- The `web_search` tool performs general open-domain knowledge search.
- The `direction` tool supports origin and destination coordinates, and may also include `waypoints`. Therefore, for multi-stop route planning, the agent may either call `direction` multiple times without waypoints, or call it once with waypoints. Evaluation should first check whether every point in the full route is covered. If all points are covered, then assess route information completeness and route rationality.
\end{PromptVerb}
\end{tcolorbox}
\captionof{figure}{Judge prompt for Open-Travel benchmark.}
\label{fig:prompt-judge-open-travel}

\begin{tcolorbox}[promptbox, title=Judge Prompt --- Open-DeepResearch]
\begin{PromptVerb}
You are a comprehensive judge for deep-research LLM agents, with expertise in information retrieval methodology, rigorous reasoning, and systematic evaluation. Given the same user query, you will compare LLM Agent A and Agent B in terms of their research paths and final answers.

The research path refers to the initial research plan and all subsequent tool-call logs. The final answer refers to the last response shown to the user after all retrieval steps. You must conduct dimension-wise quantitative evaluation, determine the overall winner, and extract reusable tactical experience from the winning trajectory.

After comparative reflection, you must provide objective scores and a winner. Then, identify the pivotal success step of the winner, and abstract it into a generalizable and reusable skill. Strictly follow the criteria, scoring rules, and JSON output format below.

## Input Format

<USER_QUERY>{original user query}</USER_QUERY>

<RETRIEVED_SKILLS>{candidate skills injected into the system prompt during trajectory generation, including skill IDs and detailed contents. Leave empty if none.}</RETRIEVED_SKILLS>

<PATH_A>{complete research path of LLM Agent A}</PATH_A>

<PATH_B>{complete research path of LLM Agent B}</PATH_B>

<ANSWER_A>{complete final answer of LLM Agent A}</ANSWER_A>

<ANSWER_B>{complete final answer of LLM Agent B}</ANSWER_B>

## Path Evaluation

### Dimensions
1. Framework: whether the first round provides a clear, progressive, and comprehensive research plan.
2. Tool Usage: whether tool calls such as `web_search` are targeted, diverse, non-redundant, and well aligned with the research steps.
3. Coverage: whether the retrieved information sufficiently covers the user need, provides a solid basis for the final answer, and avoids repetitive or redundant actions.

### Scoring Rules
- Evaluate only the research process design and tool usage, not the interpretation of retrieved information.
- Each dimension is scored from 0 to 10, where 0 means completely missing and 10 means excellent.
- The overall path score, `overall_p`, is the average of the three dimensions, rounded to one decimal place.

## Answer Evaluation

### Dimensions
1. Relevance: whether the answer fully and accurately addresses all user questions and constraints.
2. Accuracy: whether key facts, definitions, data, and conclusions are sufficiently supported, without obvious errors or contradictions.
3. Depth: whether the answer provides in-depth analysis, comparison, reasoning, and clear logical chains.
4. Clarity: whether the structure, wording, and presentation are clear, readable, and directly useful to the user.

### Scoring Rules
- Evaluate the final answer with reference to the retrieved information shown in the corresponding research path. Do not rely on external memory.
- Each dimension is scored from 0 to 10, where 0 means completely missing and 10 means excellent.
- The overall answer score, `overall_a`, is the average of the four dimensions, rounded to one decimal place.

## Overall Score and Winner

The combined score is computed as: combined_score = 0.5 * overall_p + 0.5 * overall_a

Round the combined score to one decimal place. If the two combined scores are equal, the result is `Tie`.

## Output Format

Strictly output the following JSON object and do not add any extra content:

{
  "analysis": {
    "path_comparison": "<60-150 words. Provide a comparative diagnosis of the two paths. Explain the winner's key path advantages over the loser, the loser's weaknesses, and the concrete divergence points in planning and tool usage. Specify where the loser deviated from optimal planning, such as ineffective loops or parameter errors, and how the winner avoided them.>",
    "answer_comparison": "<60-150 words. Compare the final open-domain answers and explain the winner's concrete advantages in relevance, accuracy, depth, and clarity.>"
  },
  "path_scores": {
    "Agent_A": {"framework": <0-10>, "tool_usage": <0-10>, "coverage": <0-10>, "overall_p": <0-10>},
    "Agent_B": {"framework": <0-10>, "tool_usage": <0-10>, "coverage": <0-10>, "overall_p": <0-10>}
  },
  "answer_scores": {
    "Agent_A": {"relevance": <0-10>, "accuracy": <0-10>, "depth": <0-10>, "clarity": <0-10>, "overall_a": <0-10>},
    "Agent_B": {"relevance": <0-10>, "accuracy": <0-10>, "depth": <0-10>, "clarity": <0-10>, "overall_a": <0-10>}
  },
  "combined_scores": {"Agent_A": <0-10>, "Agent_B": <0-10>},
  "winner": "<Agent_A | Agent_B | Tie>",
  "pivotal_success_step": {
    "trajectory": "<A | B>",
    "step_index": "<a list of integers or None, e.g., [3], [3,4], or None>",
    "referenced_skills": "<for each pivotal step, specify the referenced skill IDs from RETRIEVED_SKILLS that it references, e.g., [\"skill 0\"], [\"skill 0\", \"skill 1\"], or None. If a pivotal step relies only on native capability and no retrieved skill, use None at the corresponding position. If there is no pivotal step, output None.>",
    "reason": "<one concise sentence explaining what key action this step performed, why it decided the outcome, and why the referenced skill was relevant.>"
  },
  "skill": {
    "name": "<5-15 words, a concise tactical skill name>",
    "description": "<the triggering conditions, environment state, or semantic intent of the user query for applying this skill>",
    "principle": "<the core principle explaining why this strategy is logically superior>",
    "how_to_implement": "<a concrete execution guide. It may include tool-call formats, parameters, tool sequencing, and the final answer organization pattern, formulated as a standard operating procedure.>",
    "negative_constraints": "<pitfalls to avoid based on the loser's errors, covering both path-level and answer-level mistakes when applicable.>"
  }
}

## Important Requirements

- Think independently for each dimension before assigning scores to ensure fairness and objectivity.
- During scoring, completely ignore the content of <RETRIEVED_SKILLS>. Scores must be based only on the paths and final answers of Agent A and Agent B.
- All comments must be specific and traceable to the provided text. Do not introduce external information.
- The `step_index` field in `pivotal_success_step` must be a list, such as [5] or [5,6,7]. If no pivotal step exists, use None.
- The extracted skill must capture the most valuable generalizable experience, not merely summarize this instance.
- The `how_to_implement` field must include concrete tool-use rules and the final response organization paradigm.
- Focus on comparison: do not summarize A and B independently. Explain why one is better than the other in the same context, with concrete reference to step indices.
- Strictly follow the JSON template so that downstream programs can parse the output.
\end{PromptVerb}
\end{tcolorbox}
\captionof{figure}{Judge prompt for Open-DeepResearch benchmark.}
\label{fig:judge-prompt-open-deepresearch}

\begin{tcolorbox}[promptbox, title=Skill Merge Prompt]
\begin{PromptVerb}
You are an advanced skill distillation expert for optimizing complex agent tasks. Your core objective is to perform dimensionality reduction, generalization, and iterative upgrading of skill knowledge. You will receive the original complex user query, the `progressive_skills` accumulated through multiple rounds of comparative reflection, and an additional set of `existing_similar_skills`.

Your task is to refine and fuse these skills. When necessary, you should upgrade semantically similar existing skills through version evolution, and finally produce a globally optimal, highly generalizable, and reusable Golden Skill.

## Input Format

You will receive a JSON object with the following fields:

1. `query`: the original user request that triggered exploration.
2. `progressive_skills`: a list of skills accumulated across rounds.
3. `existing_similar_skills`: a set of existing skills that are semantically related to the current task.

Example input format:

{
  "query": "...",
  "progressive_skills": [
    {"round": 1, "skill": { ... }},
    {"round": 2, "skill": { ... }}
  ],
  "existing_similar_skills": "skill 1: { ... }\nskill 2: { ... }"
}

## Refinement and Fusion Principles

- Intent anchoring: deeply analyze the core pain points and implicit constraints of the query, ensuring that the generated skill addresses the essential task need.
- Trigger generalization: based on the query, abstract a single user request into a general task category or constraint combination, and place it in the `description` field.
- End-to-end SOP construction: comprehensively extract and integrate the `how_to_implement` experience from all input skills, and refine them into a complete, ordered, and declarative standard operating procedure.
- Version-evolution priority: if the distilled Golden Skill substantially overlaps with an `existing_similar_skill` in semantic intent and operational workflow, upgrade the existing skill instead of creating a duplicate. In this case, the output `name` should preserve the original title of the corresponding `existing_similar_skill` and append `_V1`, `_V2`, etc., to indicate the evolved version.
- Negative-constraint merging: comprehensively scan all `negative_constraints`. If different errors appear across rounds, such as parameter mistakes, logical loops, or missing information, merge them into a unified set of strict red-line constraints.

## Output Format

Strictly output the following JSON object and do not add any extra content:

{
  "golden_skill": {
    "name": "<5-15 words. A concise compound strategic identifier that is not limited to the specific query.>",
    "description": "<the triggering conditions, task category, prerequisites, or environment state abstracted from the query. This field is used for future vector-based semantic retrieval.>",
    "principle": "<the core theoretical principle explaining how this strategy handles typical complex queries by balancing the positive tool chain and negative constraints.>",
    "how_to_implement": "<standard execution SOP: describe steps as 1. 2. 3. Integrate tool-selection logic, core parameter-handling rules, and the structured organization of the final answer.>",
    "negative_constraints": "<golden red lines: a highly condensed set of negative constraints distilled from past mistakes, specifying actions that must never be taken to prevent the agent from falling into the same traps.>"
  },
  "fusion_rationale": "<60-100 words. Briefly explain how you used the query intent to select, fuse, and upgrade `progressive_skills` and `existing_similar_skills`, and how historical errors were distilled into `negative_constraints`.>"
}

## Important Requirements

- Never physically concatenate the text of local skills. You must perform semantic-level simplification, abstraction, and refinement.
- Strictly follow the JSON template so that downstream programs can parse the output.

## Tool Explanation

- The `web_search` tool performs general open-domain knowledge search.
\end{PromptVerb}
\end{tcolorbox}
\captionof{figure}{Skill merge prompt.}
\label{fig:skill-merge-prompt}

\clearpage

\end{document}